\documentclass[preprint,12pt]{elsarticle}

\usepackage[numbers]{natbib}
\usepackage{lineno}
\usepackage[linesnumbered,boxed,ruled,commentsnumbered]{algorithm2e}
\usepackage{xcolor}
\usepackage{hhline}
\usepackage{booktabs}
\usepackage{tabularx}
\usepackage{pifont}
\usepackage{float}
\usepackage{graphicx}
\usepackage{caption}
\usepackage{placeins}
\usepackage{mathrsfs}
\usepackage{afterpage}
\usepackage{subcaption}
\usepackage{multirow}
\usepackage{makecell}
\usepackage{amsmath}
\usepackage{rotating}
\usepackage{siunitx}
\usepackage{ulem}
\usepackage{amssymb} 
\usepackage[hyphens]{url}
\usepackage{hyperref}
\hypersetup{hidelinks}
\newcommand{\xmark}{\ding{55}}
\newcommand{\cmark}{\ding{51}}
\newcolumntype{Y}{>{\centering\arraybackslash}X}

\usepackage{tcolorbox}
\tcbuselibrary{skins}

\journal{ISPRS}

\begin{document}
	
	\begin{frontmatter}
		
	\title{GeoHeight-Bench: Towards Height-Aware Multimodal Reasoning in Remote Sensing}

	\author[1,2]{Xuran Hu}
	\author[3]{Zhitong Xiong\corref{cor1}}
	\author[2]{Zhongcheng Hong}
	\author[1]{Yifang Ban}
	\author[3]{Xiaoxiang Zhu}
	\author[2]{Wufan Zhao\corref{cor1}}

	\affiliation[1]{organization={KTH Royal Institute of Technology}}
	\affiliation[2]{organization={The Hong Kong University of Science and Technology (Guangzhou)}}
	\affiliation[3]{organization={Technical University of Munich}}
	
	\cortext[cor1]{Corresponding author: Zhitong Xiong (zhitong.xiong@tum.de), Wufan Zhao (wufanzhao@hkust-gz.edu.cn)}
		
		\begin{abstract}
		Current Large Multimodal Models (LMMs) in Earth observation are predominantly evaluated on planar optical tasks and often neglect the vertical dimension, although vertical geometric structure can be critical in applications such as disaster response and urban-morphology analysis. Progress on height-aware reasoning is also hindered by the absence of systematic evaluation: few benchmarks pair optical imagery with height products across pixel-, object-, and scene-level reasoning. To address this gap, we introduce GeoHeight-Bench, a large-scale benchmark for height-aware remote sensing understanding, together with a more challenging terrain-oriented extension, GeoHeight-Bench+. The benchmark is constructed through a scalable, VLM-driven generation pipeline that combines metadata extraction with prompt engineering, and its quality is assessed through a human-in-the-loop verification protocol. To examine whether height-aware reasoning can be learned from optical imagery, we further provide GeoHeightChat, a height-aware baseline that transfers implicit height-related geometric representations into an optical LMM. Evaluations of a broad range of closed- and open-source LMMs show that current models remain limited in their ability to reason about height information, while aligning implicit height priors improves most height-dependent tasks. However, several tasks, particularly slope reasoning and terrain-based flood-susceptibility mapping, remain largely unsolved, highlighting concrete open problems for height-aware GeoAI. Dataset and Code will be released \href{https://teriri1999.github.io/GeoHeight/}{here}. 
		\end{abstract}
		
		%%%Graphical abstract
		%\begin{graphicalabstract}
		%%\includegraphics{grabs}
		%\end{graphicalabstract}
		%
		%%%Research highlights
		%\begin{highlights}
		%\item Research highlight 1
		%\item Research highlight 2
		%\end{highlights}
		
		%% Keywords
		
	\end{frontmatter}

	\section{Introduction}
	\label{sec:intro}
	
	Large Multimodal Models (LMMs) have expanded Remote Sensing (RS) systems from task-specific prediction toward interactive visual interpretation and reasoning \cite{kuckreja2024geochat, zhang2024earthgpt, wang2026towards}. By connecting optical observations with natural language and other sensing modalities, they provide a promising interface between Earth observation data and human-oriented analysis \cite{shu2025earthmind, bazi2024rs}. Most current capabilities, however, remain planar and appearance-driven. Optical LMMs can recognize land-cover categories, localize objects, and describe spatial arrangements, but their ability to reason about vertical geometric properties—such as object height, terrain elevation, and the influence of vertical structure on spatial processes—remains limited. Such information is more than an additional semantic attribute and is critical in flood simulation, landslide susceptibility assessment, and urban morphology analysis \cite{lyu2026causal, lyu2026measuring}, where the relation between surface objects and the underlying terrain can be as important as their appearance \cite{sampson2015high, schumann2018global, jebur2014optimization}. Making this vertical dimension usable by GeoAI requires systematic evaluation, scalable multimodal supervision, and effective mechanisms for transferring height information into models that operate on optical imagery.

	Systematic evaluation remains difficult because height understanding is not a single prediction problem. RS visual question answering and grounding benchmarks mainly assess semantic recognition and localization \cite{lobry2020rsvqa, sun2022visual}, while broader evaluations emphasize general visual understanding or cross-view urban reasoning \cite{zhou2025urbench, adejumo2025vision}. The emerging VLRS-Bench incorporates information from digital surface models (DSMs), but height is one component of a broader vision--language evaluation \cite{luo2026vlrs}. A dedicated benchmark should account for the distinct physical quantities represented by Digital Terrain Models (DTMs), DSMs, and normalized Digital Surface Models (nDSMs), while evaluating height-aware reasoning across multiple spatial granularitie. Pixel-level questions retrieve local elevation, height, slope, or aspect; object-level questions associate measurements with semantic instances and compare them; scene-level questions aggregate local values into distributions or terrain descriptions; and application-oriented questions combine terrain with land cover to derive terrain-based landslide- and flood-susceptibility indicators. These capabilities require both textual answers and spatially precise masks. They should also expose cross-level consistency, since a plausible description may conflict with the predicted ordering or mask. Without a unified framework across quantities, granularities, and output forms, strong performance on one task provides limited evidence of generalizable height awareness.
	
	Existing remote sensing height products provide a quantitative foundation for constructing height-aware benchmarks. LiDAR- and InSAR-derived products, including Digital Elevation Models (DEMs), DSMs, and nDSMs, provide complementary supervision for terrain elevation and above-ground object height \cite{farr2007shuttle, wang2018lidar}. Their information is stored as georeferenced rasters rather than ready-made instructions. Constructing a question may require aligning optical and height products, coordinate or instance extraction, local or scene statistics computation, and linking those values to semantic labels. Manual authoring is accurate but costly and difficult to reproduce at scale. Fixed templates can maintain numerical accuracy but offer limited linguistic variation; whereas VLM-assisted generation can improve linguistic diversity, as demonstrated by synthetic data from geospatial instruction segmentation \cite{quenum2025lisat}, but may introduce statements inconsistent with the underlying measurements. Height-dependent questions are especially sensitive because numerical attributes, spatial relations, masks, and descriptions must remain traceable to the same source products. Geographic diversity and leakage-resistant splits are also necessary to prevent nearby tiles from inflating performance. Scalable construction therefore requires metadata-grounded instruction generation together with explicit human quality control.
	
	Inferring height information from optical imagery remains challenging, even when reliable reference data are available. Although optical imagery contains indirect geometric cues such as shadows, texture, perspective, and context, these cues do not uniquely determine absolute height from a single top-down view. Monocular height estimation shows that these cues support dense prediction, but typically decode task-specific height rasters rather than support language-based reasoning or reasoning-conditioned spatial grounding \cite{mou2018im2height, liu2020im2elevation}. Other height-aware architectures target specialized outputs such as 3D change detection or part segmentation \cite{liu2025hatformer, kareem2024paris3d}. Supplying an explicit DEM or DSM to an LMM at inference time is possible, but changes the input requirements and is impractical when the corresponding product is unavailable. Cross-modal distillation suggests that information from a training-only modality can be transferred to a model that operates without it at inference \cite{yang2025dpmamba, wang2025tpdtnet}. It remains unclear whether an implicit height representation can jointly support language generation and dense spatial grounding, or support reasoning over derived terrain properties such as slope and spatial connectivity. A height-aware baseline is therefore needed both to validate the benchmark as a learnable test and to characterize the limits of optical height inference.
	
	To address these gaps, we introduce GeoHeight-Bench and its terrain-oriented extension, GeoHeight-Bench+, forming a two-tier benchmark suite of ten tasks spanning pixel-, object-, scene-, and reasoning-level evaluation. We construct the benchmarks through metadata extraction, VLM-assisted instruction generation, and human verification. We further provide GeoHeightChat, a proof-of-concept baseline that aligns implicit geometric representations with an optical LMM. Experiments show that height alignment improves most evaluated capabilities, while slope reasoning and terrain-based flood-susceptibility mapping remain challenging.
	
	The main contributions of this paper are summarized as follows:
	
	\begin{enumerate}
		\item \textbf{A systematic benchmark for height-aware remote sensing reasoning} We introduce GeoHeight-Bench, a large-scale multimodal benchmark that systematically evaluates the vertical dimension of remote sensing understanding across pixel-, object-, and scene-level reasoning, together with a terrain-oriented extension, GeoHeight-Bench+. To our knowledge, it is among the first benchmarks to evaluate height-aware reasoning from optical imagery at this granularity and scale.
		\item \textbf{A scalable VLM-driven annotation pipeline.} To reduce the cost and expert effort required for height-aware Earth observation annotation, we design an automated, VLM-driven pipeline for metadata extraction and prompt engineering, coupled with a human-in-the-loop verification protocol, enabling scalable and quality-controlled multimodal data generation.
		\item \textbf{A height-aware remote sensing reasoning baseline.} As a proof of concept for the benchmark, we propose GeoHeightChat, an LMM baseline that transfers implicit height priors into optical representations through a two-stage alignment and instruction-tuning framework. It establishes reference results on GeoHeight-Bench and GeoHeight-Bench+, demonstrates the benefit of implicit height representations for multimodal reasoning, and exposes tasks that current height priors do not yet resolve.
	\end{enumerate}

	\section{Related Work}\label{section:A}

	\subsection{Large Multimodal Models in Remote Sensing} Early adaptations such as GeoChat \cite{kuckreja2024geochat} transferred LLaVA-style architectures to earth observation and enabled region-level reasoning through instruction tuning. To address single-modality limitations, EarthGPT \cite{zhang2024earthgpt} integrated optical, SAR, and infrared imagery, while EarthGPT-X \cite{zhang2025earthgpt} further extended flexible multi-source processing. More recent foundation models include SkySense \cite{guo2024skysense}, which uses factorized spatiotemporal encoders, and SkySense V2 \cite{zhang2025skysense}, which scales with a Mixture-of-Experts strategy. Other emerging models, including GeoRSMLLM \cite{zhang2025georsmllm}, and broader surveys \cite{lane2026genealogy, xiao2025foundation} illustrate rapid progress toward general-purpose earth observation perception. Despite these advances in visual-semantic understanding \cite{tao2025advancements}, vertical geometric structure remains underrepresented in current evaluation.

	\subsection{Reasoning Segmentation and Agents} Reasoning segmentation extends LMMs from bounding-box localization to pixel-level prediction. The LISA paradigm \cite{lai2024lisa} introduced an ``embedding as mask'' strategy, which GSVA \cite{xia2024gsva} generalized to multimodal settings. In the geospatial domain, SegEarth-R1 \cite{li2025segearth} and SegEarth-R2 \cite{xin2026segearth} incorporated spatial-attention supervision and dynamic query mechanisms to address scale variation and multi-target instructions. Similarly, LISAt \cite{quenum2025lisat} leveraged large-scale synthesized data, including GRES \cite{liu2023gres}, for fine-grained geospatial reasoning. GeoPixel \cite{shabbir2025geopixel} and OmniRIS \cite{hu2023beyond} further advanced pixel grounding and flexible prompting for complex referring expressions. These methods, however, primarily emphasize semantic attributes rather than height-related geometric properties.
	
	\subsection{Remote Sensing Benchmarks and Height Perception} Evaluating LMMs in remote sensing requires comprehensive benchmarks. UrBench \cite{zhou2025urbench} integrates satellite and street-view data for cross-view reasoning, while the Vision-Centric Benchmark \cite{adejumo2025vision} targets general visual tasks; both place limited emphasis on the vertical dimension. The most closely related concurrent effort is VLRS-Bench \cite{luo2026vlrs}, which also incorporates height products such as DSMs into a reasoning benchmark. Our work is complementary but differs in emphasis: we organize height reasoning across pixel, object, scene, and reasoning levels; distinguish normalized object height from absolute terrain elevation; and provide segmentation supervision for spatially groundable object-, scene-, and reasoning-level tasks.
	
	Estimating height itself has a long history in remote sensing. Monocular height estimation (MHE) regresses a height map directly from a single optical image: IM2HEIGHT \cite{mou2018im2height} first cast this as an end-to-end convolutional--deconvolutional mapping, and subsequent works refined it through encoder--decoder \cite{amirkolaee2019height}, adversarial \cite{ghamisi2018img2dsm}, and joint semantic--height designs \cite{liu2020im2elevation}. These methods establish that monocular cues such as shadows, texture, and perspective contain exploitable but ambiguous height information. Our work differs in output: rather than decoding a dense height raster, we align an implicit height representation that can support language and segmentation queries. Architecturally oriented works such as HATFormer \cite{liu2025hatformer} and PARIS3D \cite{kareem2024paris3d} encode height for 3D change detection or part segmentation, but do not address the multi-granular, language-driven height reasoning evaluated by GeoHeight-Bench.
	
	\subsection{Cross-Modal Feature Alignment} Aligning heterogeneous modalities is crucial for imparting geometric awareness. A prevalent strategy is cross-modal distillation, where a teacher network trained on explicit modalities guides an RGB-based student. Techniques like DPMamba \cite{yang2025dpmamba} and TPDTNet \cite{wang2025tpdtnet} demonstrate effective distillation for missing modalities. Foundation models use meta-modality attention to bridge the gap between optical and SAR data. Similarly, approaches like EarthMind \cite{shu2025earthmind} and RescueADI \cite{liu2025rescueadi} demonstrate the value of multi-granular fusion and autonomous agents in complex tasks like disaster interpretation. We adopt a dense feature alignment approach, utilizing a frozen geometric teacher to inject implicit height priors into the LMM, enabling implicit vertical reasoning during inference. This can be viewed as performing monocular height estimation \cite{mou2018im2height} in feature space rather than pixel space: the distilled prior is consumed by the language model for reasoning instead of being decoded into an explicit height map.
	
	\begin{figure*}[t]
		\centering
		\includegraphics[width=\textwidth]{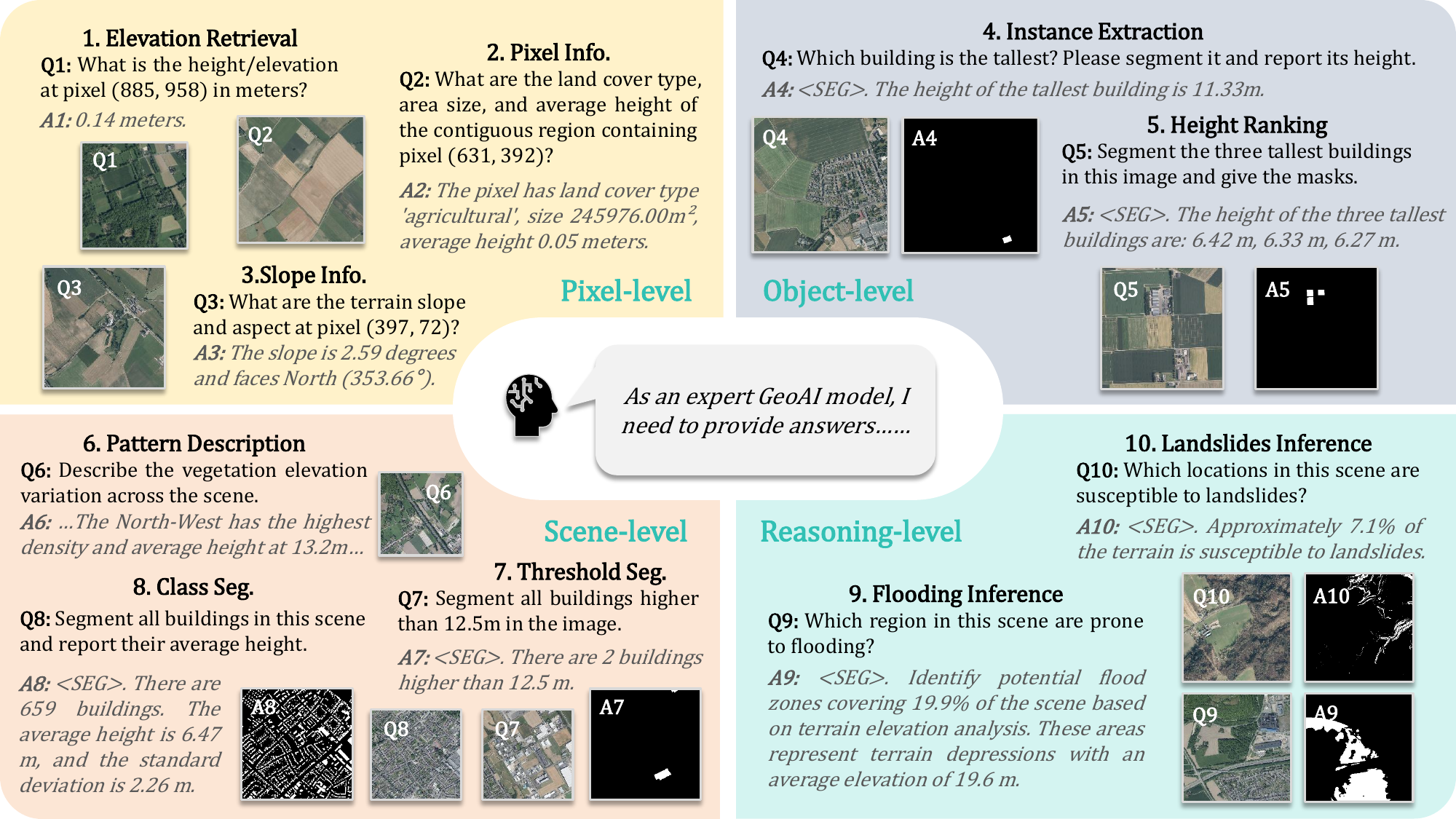}
		\caption{Overview of GeoHeight-Bench and GeoHeight-Bench+. The benchmark suite comprises 10 tasks organized into four hierarchical levels: Pixel-level retrieval, Object-level extraction, Scene-level analysis, and Reasoning-level inference.}
		\label{fig_1}
	\end{figure*}

	\section{GeoHeight-Bench}
	
	\subsection{Benchmark Tasks}
	\label{sec:benchmark_tasks}
	
	We introduce GeoHeight-Bench and GeoHeight-Bench+ to evaluate height-related reasoning in remote sensing LMMs through ten tasks organized into four levels. Figure \ref{fig_1} summarizes the task taxonomy.
	
	GeoHeight-Bench comprises tasks at three hierarchical levels: Pixel, Object, and Scene. GeoHeight-Bench+ additionally includes a Reasoning level for terrain-conditioned questions.

	\begin{enumerate}
		\item \textbf{Pixel-level} encompasses three tasks: Elevation Retrieval (ER), Pixel Information (PI), and Slope Information (SI). ER queries the height or elevation at a specified image coordinate. PI jointly evaluates semantic category, height, and spatial-scale information. SI, which is exclusive to GeoHeight-Bench+, queries terrain slope and aspect at a specified coordinate.
		
		\item \textbf{Object-level} includes two tasks: Instance Extraction (IE) and Height Ranking (HR). IE requires the model to identify an instance from a specified category and output both its height information and segmentation mask. HR requires the model to rank buildings by height and return the corresponding masks.
		
		\item \textbf{Scene-level} evaluates global information aggregation and natural language description. It contains three tasks: Pattern Description (PD), Threshold Segmentation (TS), and Class Segmentation (CS). PD describes the height distribution of land-cover categories or, in GeoHeight-Bench+, the spatial distribution of terrain slope. TS requires the model to segment objects that exceed a specified height or elevation threshold. CS requires a scene-level summary of a specified category, including global statistics and a segmentation mask.
		
		\item \textbf{Reasoning-level} consists of two tasks: Landslide Inference (LI) and Flood Inference (FI). These tasks combine elevation-derived terrain attributes with land-cover categories to produce static, terrain-based susceptibility indicators. They evaluate whether a model can localize areas that are topographically more susceptible under the benchmark's annotation rules, rather than provide complete hazard assessments.
	\end{enumerate}
	
	\begin{figure*}[t]
		\centering
		\includegraphics[width=\textwidth]{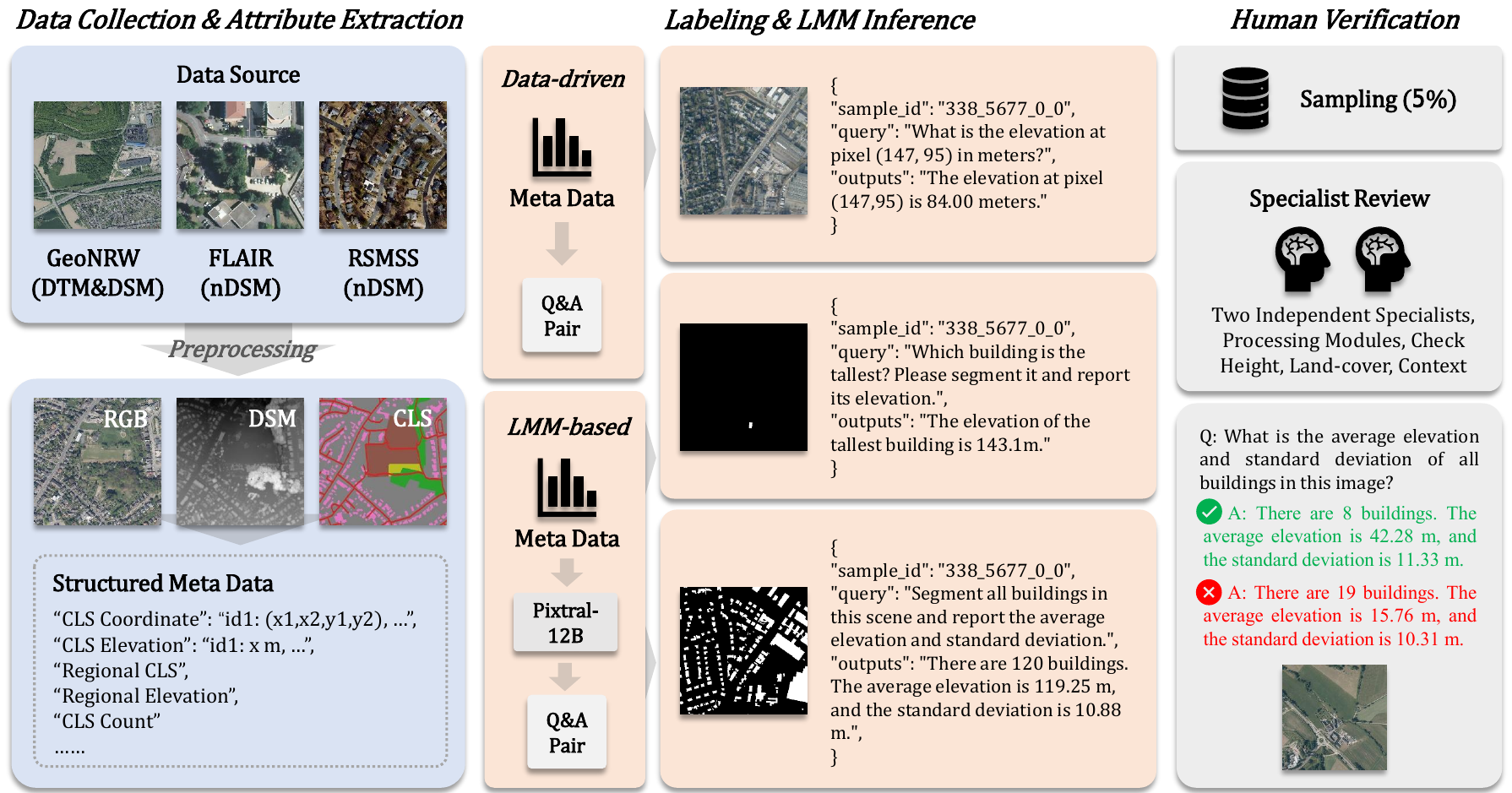}
		\caption{Pipeline of the GeoHeight-Bench generation and verification.}
		\label{fig_2}
	\end{figure*}

	\begin{table}[t]
		\centering
		\scriptsize
		\caption{Human verification with uniform task-wise sampling. A $5\%$ sample is
			single-reviewed for the overall error rate; a $1{,}000$-pair subset is
			double-annotated for inter-rater reliability and provenance analysis.}
		\label{tab:verification}
		\setlength{\tabcolsep}{5pt}
		\begin{tabular}{lcccc}
			\toprule
			Subset & \#Verified & \#Errors & Error (\%) & Acc.\ (\%) \\
			\midrule
			\multicolumn{5}{l}{\textit{Double-annotated subset}}\\
			\quad Data-driven (templated) & 375     & 0   & 0.0 & 100.0 \\
			\quad LMM-generated (Pixtral) & 625     & 26  & 4.2 & 95.8 \\
			\quad Subset overall          & 1{,}000 & 26  & 2.6 & 97.4 \\
			\midrule
			\multicolumn{5}{l}{\textit{Single-reviewed $5\%$ sample}}\\
			\quad Full $5\%$ sample        & 15{,}000 & 180 & 1.2 & 98.8 \\
			\midrule
			\multicolumn{5}{l}{\scriptsize Inter-rater (subset): raw agr.\ $98.5\%$, $\kappa=0.59$, Gwet's AC1 $=0.98$.}\\
			\bottomrule
		\end{tabular}
	\end{table}

	\subsection{Benchmark Pipeline}

	The construction of GeoHeight-Bench follows three stages: attribute extraction, prompt design and LMM inference, and human verification. Each stage refines the data representation to improve completeness, consistency, and semantic accuracy across modalities. Figure \ref{fig_2} summarizes the workflow.
	
	\vspace{5pt}
	
	%https://www.opengeodata.nrw.de/produkte/
	
	\noindent \textbf{Data Collection and Attribute Extraction.} Large vision-language models remain limited in their ability to recover fine-grained spatial cues and quantitative elevation patterns directly from remote sensing imagery. We therefore construct structured metadata from GeoNRW \cite{s5xq-b822-20}, RSMSS \cite{xiong2022parse}, and FLAIR \cite{garioud2026flair}, together with official NRW DTM products. Land-cover maps and connected-component analysis are used to identify coherent regions. For each region, we compute descriptive statistics, including mean height or elevation, and combine them with categorical metadata. Task-specific filters remove redundant information. For information-retrieval tasks that do not require language-model reasoning, answers are derived directly from the metadata and inserted into predefined templates.

	\vspace{5pt}
	
	\noindent \textbf{Prompt Design and LMM Inference.} Based on the extracted metadata, we construct a two-tier prompt schema inspired by multimodal instruction tuning \cite{dai2023instructblip, touvron2023llama}. The system prompt specifies global output constraints, while the user prompt encodes task-specific requests such as identifying the highest object, comparing elevation gradients, or reasoning about object--terrain interactions.

	We adopt Pixtral-12B \cite{agrawal2024pixtral} as the vision-language reasoning backbone. Pixtral combines a dedicated vision encoder with a multimodal language decoder and supports variable image resolutions and aspect ratios. During QA generation, we apply iterative consistency checking: the model is prompted with multiple linguistic formulations, and inconsistent responses are filtered through semantic matching. The retained answers are post-processed to remove uncertain or underspecified phrases (e.g., ``possibly'' and ``it seems'') and normalized to a consistent response structure.
	
	\vspace{5pt}
	
	%	\noindent \textbf{Human Verification:} While automated inference ensures scalability, high-quality benchmarks in scientific domains demand rigorous validation \cite{lobry2020rsvqa, wang2024skyscript}. We therefore integrate a human-in-the-loop verification stage to guarantee the accuracy and interpretability of the generated annotations. We extract a 2\% sample of the total data for validation, where each QA pair is independently reviewed by at least two remote sensing specialists, who verify the correctness of height descriptions, land-cover interpretations, and contextual reasoning. Additionally, we employ random stratified sampling for secondary quality auditing to estimate annotation precision and inter-rater agreement. After this multi-round verification, no major inconsistencies or conceptual errors were found, confirming the benchmark's reliability and readiness for large-scale evaluation.
	
	\noindent \textbf{Human Verification.} To validate annotation quality, we apply a two-tier protocol with uniform task-wise sampling. First, a task-balanced random sample of $15{,}000$ pairs (about $5\%$ of the corpus), with equal numbers drawn from each benchmark task, is single-reviewed by remote sensing specialists, yielding an overall accuracy of $98.8\%$ (error rate $1.2\%$, $180$ pairs). Second, to assess reliability and isolate the model-generated content, a task-balanced subset of $1{,}000$ pairs, again sampled uniformly across tasks, is double-annotated by two independent specialists---a pair is deemed correct only if it jointly passes height, land-cover, and contextual consistency, with the $15$ disagreements settled by a third expert. On this subset the annotators reach $98.5\%$ raw agreement; the modest Cohen's $\kappa=0.59$ is a known artefact of the kappa paradox under high prevalence~\cite{feinstein1990high}, while the prevalence-robust Gwet's $\text{AC1}=0.98$~\cite{gwet2008computing} confirms almost-perfect agreement. Crucially, all errors originate from the LMM-generated subset ($4.2\%$, $95\%$ Wilson CI $[2.9\%,6.0\%]$), whereas the metadata-templated subset is error-free (Table~\ref{tab:verification}); this directly addresses the concern that a model-generated ground truth might propagate backbone errors. All flagged pairs were corrected against source metadata or removed, so the released benchmark reflects the post-verification state.

	\subsection{Details of GeoHeight-Bench $(+)$}
	
	\begin{table}[t]
		\centering
		\scriptsize
		\renewcommand{\arraystretch}{1.15}
		\caption{Source datasets used in GeoHeight-Bench $(+)$.}
		\label{tab:datasets}
		\begin{tabular}{llcc}
			\toprule
			\textbf{Dataset} & \textbf{Region} & \textbf{Res.} & \textbf{Height Data} \\
			\midrule
			\textbf{GeoNRW} & Germany & 1.0 m & DSM, DTM \\
			\textbf{FLAIR}  & France  & 0.2 m & nDSM \\
			\textbf{RSMSS}  & USA     & $\sim$0.4 m & nDSM \\
			\bottomrule
		\end{tabular}
	\end{table}
	
	%	\begin{figure}[t]
		%		\centering
		%		\includegraphics[width=0.75\columnwidth]{figures/task_sunburst.pdf}
		%		\caption{Composition of GeoHeight-Bench $(+)$ by hierarchy level (inner ring,
			%			with level shares) and task (outer ring, with per-task pair counts), wedge
			%			angles proportional to the number of instruction-tuning pairs. Hatched wedges
			%			(SI, LI, FI) are terrain-dependent tasks instantiated only in GeoHeight+.}
		%		\label{fig:task_sunburst}
		%	\end{figure}

	GeoHeight-Bench is built on three publicly available, height-annotated remote sensing sources, GeoNRW~\cite{s5xq-b822-20}, FLAIR~\cite{garioud2026flair}, and RSMSS~\cite{xiong2022parse}, summarized in Table~\ref{tab:datasets}. The three sources span Germany, France, and the USA at spatial resolutions from 0.2\,m to 1.0\,m, and together cover the full range of height products required by our tasks: Digital Surface Models (DSM), Digital Terrain Models (DTM), and normalized DSMs (nDSM). This diversity is deliberate: the DTM-bearing GeoNRW supplies the bare-earth elevation needed for slope reasoning and terrain-based flood- and landslide-susceptibility mapping, whereas the higher-resolution nDSM sources (FLAIR and RSMSS) sharpen object-level height discrimination. The semantic categories supported by GeoHeight-Bench are aligned with the source-specific label ontologies of GeoNRW, RSMSS, and FLAIR. Each question and reference mask uses only categories defined for its corresponding source dataset. The reported benchmark setting is closed-set and is evaluated on the predefined categories. The pretrained CLIP vision encoder retains some open-vocabulary recognition capability.
	
	To prevent spatial and semantic leakage, we partition all source imagery at the original-tile level before QA generation. All QA pairs derived from the same source tile, including those belonging to different task categories, are assigned exclusively to one split. Consequently, no source tile contributes samples to more than one of the training, validation, and test sets.
	
\begin{figure*}[t]
		\centering
		\begin{subfigure}{0.49\textwidth}
			\centering
			\includegraphics[width=\textwidth]{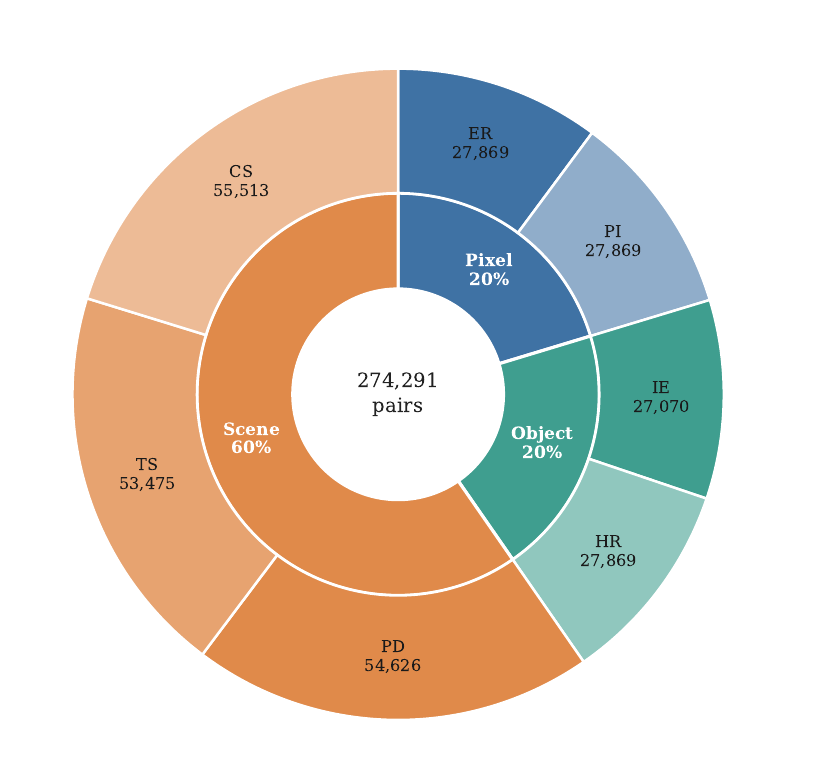}
			\caption{GeoHeight-Bench (274{,}291 pairs).}
			\label{fig:sunburst_geo}
		\end{subfigure}
		\hfill
		\begin{subfigure}{0.49\textwidth}
			\centering
			\includegraphics[width=\textwidth]{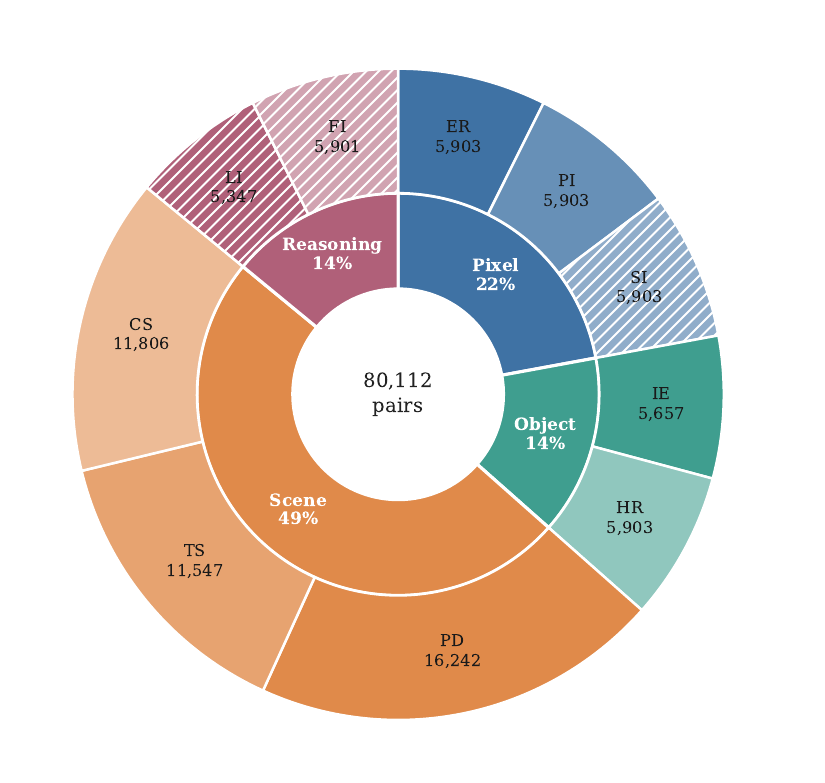}
			\caption{GeoHeight-Bench+ (80{,}112 pairs).}
			\label{fig:sunburst_geoplus}
		\end{subfigure}
		\caption{Composition of each benchmark by hierarchy level (inner ring) and
			task (outer ring); wedge angles proportional to the number of
			instruction-tuning pairs. GeoHeight-Bench covers Pixel/Object/Scene levels,
			while GeoHeight-Bench+ adds the Reasoning level and the terrain-dependent
			tasks SI, LI, and FI (hatched).}
		\label{fig:sunbursts}
	\end{figure*}

	In total, GeoHeight-Bench comprises 274{,}291 instruction-tuning pairs across pixel-, object-, and scene-level tasks, while the terrain-centric GeoHeight-Bench+ contributes a further 80{,}112 pairs and introduces the terrain-dependent tasks SI, LI, and FI, instantiated only on DTM-bearing sources. Figure~\ref{fig:sunbursts} shows the per-task composition of both benchmarks.
	
	To characterize the semantic content of the generated instructions beyond raw counts, we analyze the lexical distribution of the QA pairs: user prompts are dominated by spatial and geometric reasoning terms (e.g., ``tallest,'' ``elevation,'' ``distribution pattern''), whereas the ground-truth answers are anchored in absolute physical scales (e.g., ``meters,'' ``average height,'' ``standard deviation''). This contrast reflects the benchmark's intended shift of evaluation focus from planar visual textures toward the vertical dimension.

	\section{GeoHeightChat}
	\label{sec:methodology} % Changed from \label{section:B} for standard naming conventions
	
	This section presents GeoHeightChat, a height-aware LMM baseline that performs question answering and segmentation through implicit geospatial understanding. The core of our approach is to bridge the modality gap between optical remote sensing imagery and geospatial information (e.g., elevation, terrain), so that inference can rely solely on optical images. Specifically, we employ a two-stage training paradigm consisting of: (1) Cross-Modal Geo-Alignment, and (2) Geo-Aware Instruction Tuning. Because object height (nDSM) and absolute terrain elevation (DEM/DTM) are physically distinct quantities, we instantiate two variants that share the same architecture and training procedure but are trained on the corresponding height products: one on the nDSM sources for GeoHeight, and one on the DEM/DTM source for GeoHeight+. For brevity we describe a single variant below, as the pipeline is identical.
	
	\begin{figure*}[t]
		\centering
		\includegraphics[width=\textwidth]{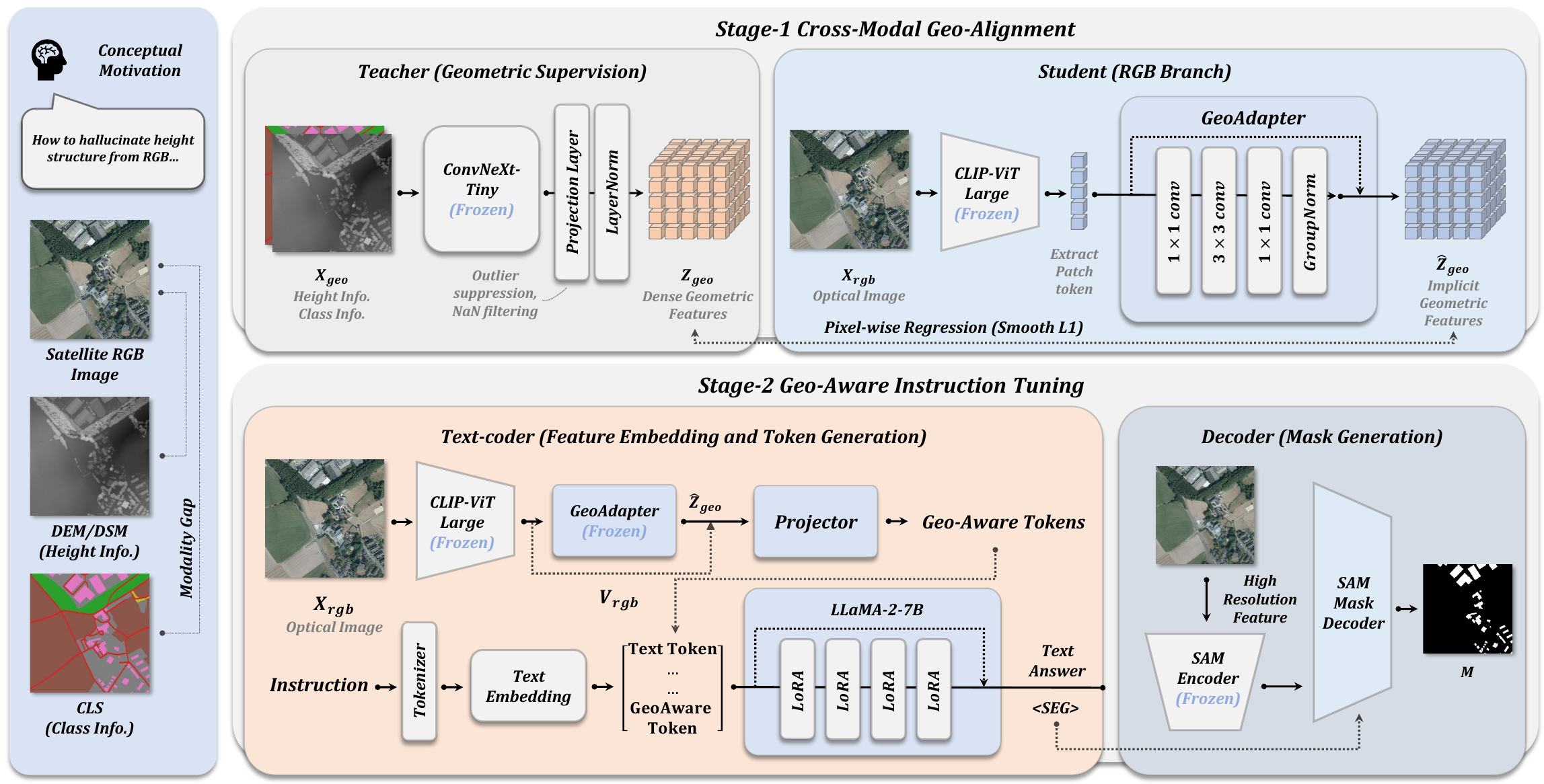}
		\caption{The proposed GeoHeightChat framework comprises two training stages: Cross-Modal Geo-Alignment and Geo-Aware Instruction Tuning.}
		\label{fig_3}
	\end{figure*} 
	
	%-------------------------------------------------------------------------
	\subsection{Problem Formulation}
	\label{sec:problem_formulation}
	
	Let $\mathcal{D} = \{(X_{rgb}^i, X_{geo}^i, T_{inst}^i, T_{gt}^i, M_{gt}^i)\}_{i=1}^N$ denote a dataset containing $N$ samples. Here, $X_{rgb}^i \in \mathbb{R}^{H \times W \times 3}$ is a high-resolution optical image, and $X_{geo}^i \in \mathbb{R}^{H \times W \times C_{geo}}$ contains two or three channels encoding height/elevation and semantic information. $T_{inst}^i$ is the language instruction, $T_{gt}^i$ is the corresponding ground-truth answer, and $M_{gt}^i \in \{0, 1\}^{H \times W}$ is the binary ground-truth segmentation mask.
	
	Our goal is to learn a mapping function $\mathcal{F}_{\theta}: (X_{rgb}, T_{inst}) \to (\hat{M}, \hat{T})$, where $\hat{M}$ and $\hat{T}$ are the predicted mask and answer. Crucially, while $X_{geo}$ is only available during the training of the alignment module (Stage 1), it is unavailable during the inference process. The model must implicitly infer the geometric features $\mathcal{Z}_{geo}$ solely from $X_{rgb}$ to support complex reasoning tasks that typically require explicit height information.
	
	%-------------------------------------------------------------------------
	\subsection{Architecture Overview}
	\label{sec:architecture}
	
	The GeoHeightChat architecture is built upon LISA-7B \cite{lai2024lisa}, extended with our GeoEncoder and GeoAdapter. The framework consists of five main components:
	
	\begin{enumerate}
		\item \textbf{Vision Backbone ($\mathcal{V}$):} We employ the pretrained CLIP ViT-Large \cite{radford2021learning} as the visual encoder. Given an RGB image $X_{rgb}$, it extracts semantic visual features $\mathbf{V}_{rgb} = \mathcal{V}(X_{rgb})$, where $\mathbf{V}_{rgb} \in \mathbb{R}^{L \times D_v}$.
		
		\item \textbf{GeoEncoder ($\mathcal{G}$):} Serving as the geometric teacher, we utilize a frozen ConvNeXt-Tiny backbone \cite{liu2022convnet} initialized with ImageNet weights. It processes the auxiliary geospatial data $X_{geo}$ (e.g., Elevation and Semantic Maps) to extract dense structural embeddings $\mathbf{Z}_{real} = \mathcal{G}(X_{geo})$. This module provides the explicit geometric supervision signals for aligning the RGB features.
		
		\item \textbf{GeoAdapter ($\mathcal{A}$):} A lightweight, trainable module designed to project optical visual features into a geometry-aware latent space. It transforms $\mathbf{V}_{rgb}$ into implicit geometric features $\hat{\mathbf{Z}}_{geo} = \mathcal{A}(\mathbf{V}_{rgb})$. Structurally, the GeoAdapter consists of a sequence of ResNet Bottleneck blocks to preserve high-frequency details essential for dense prediction tasks.
		
		\item \textbf{Large Language Model ($\mathcal{L}$):} Following LISA-7B, we utilize Llama-2-7B \cite{touvron2023llama} as the reasoning core. To enable efficient fine-tuning, we inject LoRA modules \cite{hu2022lora} into the query ($W_q$) and value ($W_v$) projections of the Transformer attention blocks. The LLM processes the multimodal sequence of text embeddings and visual embeddings to generate a response text and a special token \texttt{[SEG]}.
		
		\item \textbf{Visual Encoder ($\mathcal{E}_{v}$) and Mask Decoder ($\mathcal{D}_{mask}$):} Following LISA, we adopt the Segment Anything Model (SAM) \cite{kirillov2023segment} encoder and decoder. The visual encoder ($\mathcal{E}_{v}$) captures fine-grained spatial features, and the decoder ($\mathcal{D}_{mask}$) takes the embedding of the \texttt{[SEG]} token as a prompt (along with the visual feature maps) to generate the binary mask $\hat{M}$.
	\end{enumerate}
	
	\subsection{Cross-Modal Geo-Alignment}
	\label{sec:stage1}
	
	The primary goal of this stage is to train a lightweight adapter that can translate visual features from a frozen CLIP encoder into a geometry-aware feature space defined by a geometric teacher. Unlike contrastive approaches that learn global semantic alignment, we adopt a dense feature distillation paradigm to rigorously preserve both the semantic direction and the activation magnitude of the geospatial features for precise dense prediction.
	
	\vspace{5pt}
	
	\noindent \textbf{The Teacher: GeoEncoder.} We employ a ConvNeXt-Tiny backbone initialized with ImageNet weights as the geometric teacher $\mathcal{G}$. The input to the teacher consists strictly of geospatial data $X_{geo}$. To adapt the pre-trained backbone to 2/3-channel geo-data, we append a zero-filled channel to $X_{geo}$ before processing.
	The teacher extracts hierarchical geometric features, which are then projected and normalized to form the regression targets:
	\begin{equation}
		\mathbf{Z}_{real} = \text{LayerNorm}(\text{Proj}(\mathcal{G}(X_{geo})))
	\end{equation}
	where $\mathbf{Z}_{real} \in \mathbb{R}^{L \times D}$ represents dense geometric embeddings at $L$ spatial locations. To ensure numerical stability during regression, we suppress outliers in $\mathbf{Z}_{real}$ by clamping values to a robust range.
	
	\vspace{5pt}
	
	\noindent \textbf{The Student: GeoAdapter.} The student branch processes the RGB image $X_{rgb}$ using a frozen CLIP ViT-Large encoder. We extract the patch tokens $\mathbf{V}_{rgb}$ from the penultimate layer. The core component, GeoAdapter ($\mathcal{A}$), is designed as a stack of Bottleneck Residual Blocks. Each block consists of:
	\begin{equation}
		\text{Block}(\mathbf{x}) = \mathbf{x} + \gamma \cdot \mathcal{F}_{conv}(\text{GroupNorm}(\mathbf{x}))
	\end{equation}
	where $\mathcal{F}_{conv}$ denotes the bottleneck sequence. Crucially, we employ a Zero-Initialization strategy for the learnable scaling factor $\gamma$ (initialized to 0). This ensures that the GeoAdapter acts as an identity function at the start of training, preserving the semantic integrity of CLIP features while gradually learning geometric shifts.
	
	\vspace{5pt}
	
	\noindent \textbf{Distillation Objective: }We formulate the alignment as a regression problem. The objective is to minimize the reconstruction error between the predicted features $\hat{\mathbf{Z}}_{geo} = \mathcal{A}(\mathbf{V}_{rgb})$ and the teacher's features $\mathbf{Z}_{real}$. We utilize the Smooth L1 Loss \cite{girshick2015fast} to be robust against outliers in the geospatial data:
	\begin{equation}
		\mathcal{L}_{align} = \frac{1}{L} \sum_{i=1}^{L} \text{SmoothL1}(\hat{\mathbf{z}}_i, \mathbf{z}_i^{real})
	\end{equation}
	where $\hat{\mathbf{z}}_i$ and $\mathbf{z}_i^{real}$ denote the feature vectors at spatial location $i$. This forces the adapter to infer explicit height and semantic structures solely from RGB texture cues.

	\subsection{Geo-Aware Instruction Tuning}
	\label{sec:stage2}
	
	In the second stage, we integrate the pre-trained GeoAdapter into the LISA framework for joint question answering and segmentation.
	
	\vspace{5pt}
	
	\noindent \textbf{Geo-Aware Adapter Integration: }To preserve the learned geometric alignment, we freeze both the CLIP encoder and the GeoAdapter. The aligned features $\hat{\mathbf{Z}}_{geo}$ act as a ``structural prior'' and are fused with the standard visual features $\mathbf{V}_{rgb}$ via the projector $\mathcal{P}$ to form Geo-Aware tokens $H_v$:
	\begin{equation}
		H_v = \mathcal{P}(\mathbf{V}_{rgb} + \lambda \cdot \hat{\mathbf{Z}}_{geo})
	\end{equation}
	where $\lambda$ is a learnable scalar balancing texture and geometry. These tokens are concatenated with text instructions to fine-tune the LLM. Following LISA, we employ LoRA for efficient fine-tuning of the LLM and update the SAM mask decoder, while keeping other components frozen.
	
	\vspace{5pt}
	
	\noindent \textbf{Geometric Reasoning Inference: }During inference, the model operates in a purely RGB mode. The input $X_{rgb}$ passes through the frozen GeoAdapter, which injects the learned structural priors (e.g., relative height) into the LLM. This enables GeoHeightChat to interpret height-conditioned instructions such as "Segment the building with the highest elevation" by leveraging latent geometric features, without explicit height data at inference time.
	
	\vspace{5pt}
	
	\noindent \textbf{Training Objectives.} The total objective $\mathcal{L}_{total}$ is a weighted sum of the autoregressive text-generation loss and the two segmentation losses:
	\begin{equation}
		\mathcal{L}_{total} = \lambda_{txt} \mathcal{L}_{txt} + \lambda_{bce} \mathcal{L}_{bce} + \lambda_{dice} \mathcal{L}_{dice}
	\end{equation}
	where $\mathcal{L}_{txt}$ is the cross-entropy loss for next-token prediction, and $\mathcal{L}_{bce}$ and $\mathcal{L}_{dice}$ supervise the SAM decoder predictions against the ground-truth masks $M_{gt}$ \cite{milletari2016v}. Optimizing this objective allows the LoRA modules and mask decoder to use the geometric cues supplied by the GeoAdapter for text and mask prediction.

	\begin{figure*}[t]
		\centering
		\includegraphics[width=\textwidth]{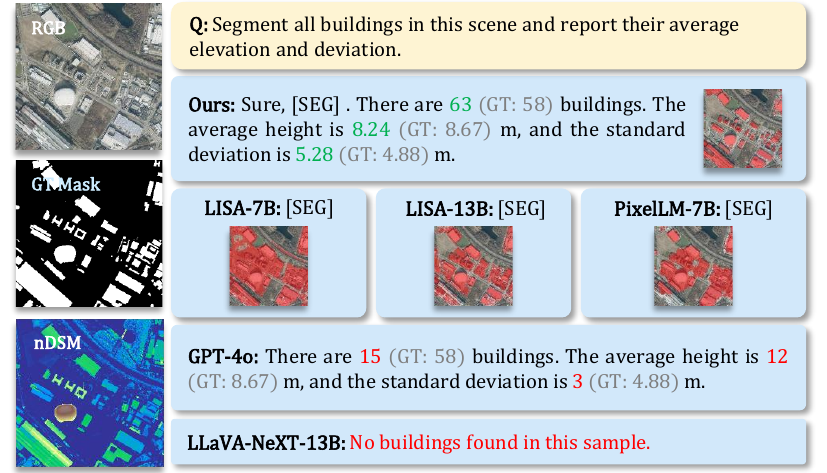}
		\caption{Comparison between GeoHeightChat and LMMs.}
		\label{fig_4}
	\end{figure*} 
	
	\begin{figure}[p]
		\centering
		\includegraphics[width=0.85\textwidth]{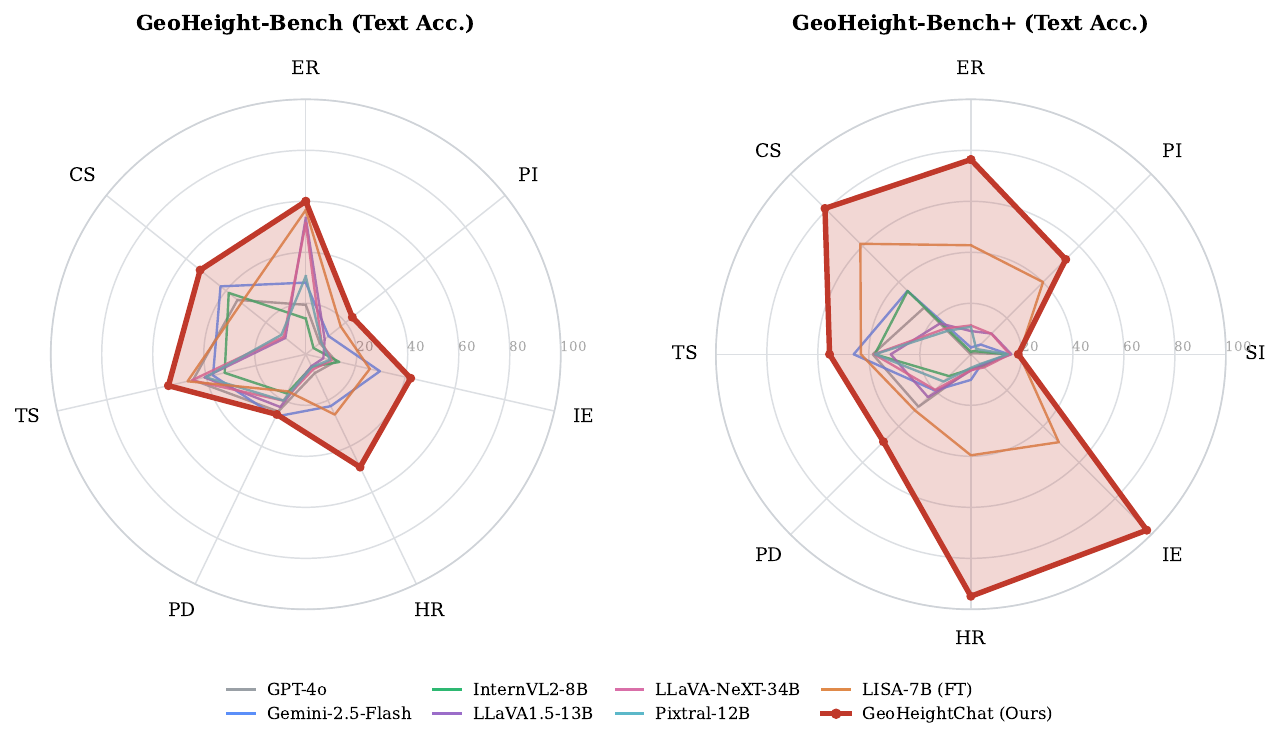}
		\caption{Per-task text accuracy (\%) on GeoHeight-Bench (left) and
			GeoHeight-Bench+ (right). GeoHeightChat (filled, red) is compared against
			closed-source (GPT-4o, Gemini-2.5-Flash) and open-source (InternVL2,
			LLaVA-1.5/NeXT, Pixtral) LMMs, together with the geo-augmented LISA-7B (FT)
			baseline.}
		\label{fig:radar_text}
	\end{figure}
	
	\begin{figure}[p]
		\centering
		\includegraphics[width=0.85\textwidth]{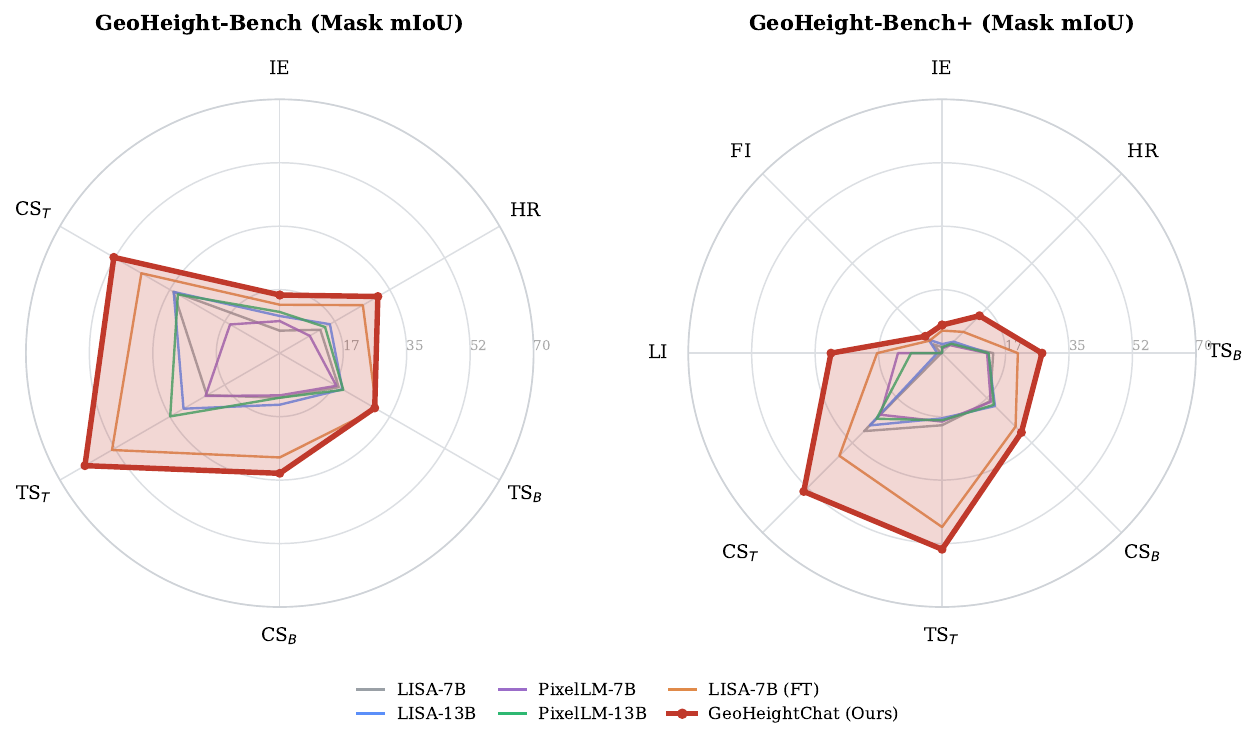}
		\caption{Per-task mask generation (mIoU) on GeoHeight-Bench (left) and
			GeoHeight-Bench+ (right), with scene-level scores split into building
			($_B$) and tree ($_T$) targets. GeoHeightChat (filled, red) is compared
			against the LISA and PixelLM segmentation baselines.}
		\label{fig:radar_mask}
	\end{figure}

	\section{Experimental Results}
	
	In this section, we comprehensively evaluate various LMMs on GeoHeight-Bench to compare the performance of existing models with our proposed GeoHeightChat on height-aware and spatial reasoning tasks. We separate two physically distinct height conventions: object height above the local ground (nDSM, used in GeoHeight) and absolute terrain elevation (DEM/DTM, used in GeoHeight+).

	\subsection{Experiment Setting}

	\noindent \textbf{Compared Models.} To ensure a comprehensive evaluation, we select 2 closed-source models and 11 open-source LMMs for comparison on GeoHeight-Bench and GeoHeight-Bench+, along with 4 multimodal reasoning segmentation models. The closed-source models include GPT-4o \cite{openai2024gpt4o} and Gemini-2.5-Flash \cite{comanici2025gemini}. The open-source large multimodal models encompass the InternVL series \cite{chen2024internvl}, LLaVA series \cite{liu2024improved}, LLaVA-NeXT series \cite{liu2024llavanext}, as well as Ministral and Pixtral models \cite{liu2026ministral, agrawal2024pixtral}. For the mask generation task, we benchmark against the LISA \cite{lai2024lisa} and PixelLM series \cite{ren2024pixellm}. Additionally, to validate the effectiveness of explicitly introducing geospatial information, we fine-tune the LISA-7B model (denoted as LISA-7B FT) by concatenating optical, height, and categorical information into a three-channel input, serving as another baseline.
	
	\vspace{5pt}
	
	\noindent \textbf{Evaluation Strategy.} We establish a standardized response-processing pipeline for different task settings. For template-based QA, we use Qwen2.5-1.5B \cite{qwen2024qwen25} to extract structured answers and apply a relative-error threshold of 20\% to continuous predictions. For open-ended questions, we use Qwen2.5-7B as a judge model to assess semantic and factual consistency \cite{zheng2023judging,qwen2024qwen25}. For mask generation, we report mean Intersection over Union (mIoU) and cumulative Intersection over Union (cIoU). Detailed evaluation protocols and prompts are provided in the supplementary material.

	\begin{table*}[p]
		\centering
		\caption{Quantitative results for text generation and spatial reasoning on GeoHeight-Bench. Scores are reported as percentages. The best and second-best results are shown in \textbf{bold} and \underline{underlined}, respectively.}\label{tab_1}
		\scriptsize
		%	\tiny
		\begin{tabular*}{\textwidth}{@{\extracolsep{\fill}} l cc cc ccc c @{}}
			\toprule
			\multirow{2}{*}{\textbf{Model}} & \multicolumn{2}{c}{\textbf{Pix-level}} & \multicolumn{2}{c}{\textbf{Obj-level}} & \multicolumn{3}{c}{\textbf{Sce-level}} & \multirow{2}{*}{\textbf{Overall}} \\
			\cmidrule(lr){2-3} \cmidrule(lr){4-5} \cmidrule(lr){6-8}
			& \textbf{ER} & \textbf{PI} & \textbf{IE} & \textbf{HR} & \textbf{PD} & \textbf{TS} & \textbf{CS} & \\
			\midrule
			
			\multicolumn{9}{l}{\textcolor{gray}{\textit{Closed-source LMMs}}} \\
			\quad GPT-4o & 19.40 & 7.02 & 12.20 & 8.18 & 24.89 & 45.34 & 34.23 & 21.61 \\
			\quad Gemini-2.5-Flash & 28.11 & 11.45 & \underline{29.82} & 22.57 & \underline{27.11} & 37.30 & \underline{42.74} & 28.45 \\
			\midrule
			
			\multicolumn{9}{l}{\textcolor{gray}{\textit{Open-source LMMs}}} \\
			\quad InternVL2-2B & 17.78 & 5.77 & 18.80 & 5.27 & 15.75 & 39.19 & 10.94 & 16.22 \\
			\quad InternVL2-4B & 12.22 & 10.27 & 22.00 & 7.75 & 21.90 & 46.44 & 34.04 & 22.09 \\
			\quad InternVL2-8B & 14.01 & 3.90 & 13.51 & 5.41 & 17.06 & 32.62 & 38.63 & 17.88 \\
			\quad LLaVA1.5-7B & 36.72 & 9.00 & 7.83 & 5.41 & \textbf{27.59} & 44.73 & 10.03 & 20.19 \\
			\quad LLaVA1.5-13B & 53.46 & 9.48 & 7.07 & 5.14 & 22.96 & 40.76 & 10.25 & 21.30 \\
			\quad LLaVA-NeXT-7B-Mistral & 22.78 & 1.97 & 7.18 & 5.78 & 19.33 & 44.99 & 10.76 & 16.11 \\
			\quad LLaVA-NeXT-7B-Vicuna & 34.07 & 9.16 & 9.63 & 5.83 & 18.06 & 44.54 & 10.12 & 18.78 \\
			\quad LLaVA-NeXT-13B & 55.64 & 9.48 & 9.27 & 7.33 & 20.40 & 44.98 & 10.42 & 22.50 \\
			\quad LLaVA-NeXT-34B & 51.37 & 7.85 & 10.43 & 6.53 & 20.17 & 45.84 & 11.00 & 21.88 \\
			\quad Ministral3-8B & 17.67 & 6.17 & 5.28 & 4.05 & 15.77 & 33.14 & 8.15 & 12.89 \\
			\quad Pixtral-12B & 30.60 & 7.77 & 9.71 & 5.52 & 20.17 & 40.30 & 12.23 & 18.04 \\
			\midrule
			\multicolumn{9}{l}{\textcolor{gray}{\textit{Height-aware LMMs}}} \\
			\quad LISA-7B (FT) & \underline{56.58} & \underline{17.53} & 25.75 & \underline{26.25} & 16.12 & \underline{47.54} & 32.31 & \underline{31.73} \\
			\quad GeoHeightChat & \textbf{59.98} & \textbf{23.36} & \textbf{42.19} & \textbf{49.09} & 26.16 & \textbf{55.24} & \textbf{52.97} & \textbf{44.14} \\
			\bottomrule
		\end{tabular*}
	\end{table*}
	
	\begin{table*}[p]
		\centering
		\caption{Quantitative results for text generation on GeoHeight-Bench+. Scores are reported as percentages.}\label{tab_2}
		
		\resizebox{\textwidth}{!}{
			\begin{tabular}{l ccc cc ccc c}
				\toprule
				\multirow{2}{*}{\textbf{Model}} & \multicolumn{3}{c}{\textbf{Pix-level}} & \multicolumn{2}{c}{\textbf{Obj-level}} & \multicolumn{3}{c}{\textbf{Sce-level}} & \multirow{2}{*}{\textbf{Overall}} \\
				\cmidrule(lr){2-4} \cmidrule(lr){5-6} \cmidrule(lr){7-9}
				& \textbf{ER} & \textbf{PI} & \textbf{SI} & \textbf{IE} & \textbf{HR} & \textbf{PD} & \textbf{TS} & \textbf{CS} & \\
				\midrule
				
				\multicolumn{10}{l}{\textcolor{gray}{\textit{Closed-source LMMs}}} \\
				\quad GPT-4o & 0.00 & 0.68 & 15.03 & 6.01 & 5.74 & 29.05 & 38.51 & 25.90 & 15.12 \\
				\quad Gemini-2.5-Flash & 2.70 & 5.41 & 14.86 & 5.71 & 10.03 & 20.04 & \underline{45.95} & 35.14 & 17.48 \\
				\midrule
				
				\multicolumn{10}{l}{\textcolor{gray}{\textit{Open-source LMMs}}} \\
				\quad InternVL2-2B & 0.68 & 0.56 & 14.18 & 7.07 & 6.41 & 15.52 & 37.33 & 13.51 & 11.91 \\
				\quad InternVL2-4B & 5.41 & 16.58 & 10.14 & 7.42 & 8.20 & 20.29 & 38.34 & 33.61 & 17.50 \\
				\quad InternVL2-8B & 1.01 & 1.80 & 14.53 & 6.71 & 5.74 & 12.09 & 37.67 & 35.08 & 14.33 \\
				\quad LLaVA1.5-7B & 5.07 & 10.26 & 16.05 & 6.36 & 5.78 & 23.91 & 38.01 & 11.66 & 14.64 \\
				\quad LLaVA1.5-13B & 9.12 & 11.50 & 15.88 & 6.01 & 6.35 & 23.78 & 31.42 & 17.12 & 15.15 \\
				\quad LLaVA-NeXT-7B-Mistral & 2.27 & 0.00 & 14.86 & 6.07 & 4.62 & 16.63 & 38.01 & 11.82 & 11.79 \\
				\quad LLaVA-NeXT-7B-Vicuna & 7.43 & 10.59 & 16.39 & 7.77 & 5.06 & 18.08 & 37.82 & 11.88 & 14.38 \\
				\quad LLaVA-NeXT-13B & 7.09 & 12.62 & 15.54 & 6.71 & 5.93 & 12.99 & 38.07 & 12.44 & 13.92 \\
				\quad LLaVA-NeXT-34B & 11.32 & 11.38 & 15.24 & 7.27 & 6.20 & 19.76 & 37.91 & 14.26 & 15.42 \\
				\quad Ministral3-8B & 8.60 & 2.01 & 12.77 & 5.26 & 4.87 & 13.64 & 37.75 & 11.73 & 12.08 \\
				\quad Pixtral-12B & 10.75 & 2.96 & 14.60 & 5.34 & 5.29 & 15.11 & 37.60 & 12.98 & 13.08 \\
				\midrule
				\multicolumn{10}{l}{\textcolor{gray}{\textit{Height-aware LMMs}}} \\
				\quad LISA-7B (FT) & \underline{42.76} & \underline{40.03} & \textbf{18.96} & \underline{48.74} & \underline{39.60} & \underline{31.06} & 43.11 & \underline{61.29} & \underline{40.69} \\
				\quad GeoHeightChat & \textbf{76.35} & \textbf{52.62} & \underline{18.58} & \textbf{97.53} & \textbf{94.87} & \textbf{48.48} & \textbf{55.41} & \textbf{80.91} & \textbf{65.59} \\
				\bottomrule
			\end{tabular}
		}
	\end{table*}

	\begin{table*}[p]
		\centering
		\caption{Fine-grained mask generation on GeoHeight-Bench, evaluated using mIoU (\%) and cIoU (\%). Scene-level results are reported separately for buildings (B) and trees (T).}\label{tab_3}
		\scriptsize
		\begin{tabular*}{\textwidth}{@{\extracolsep{\fill}} l cc cc cc c@{}}
			\toprule
			\multirow{2}{*}{\textbf{Model}} & \multicolumn{2}{c}{\textbf{Obj-level}} & \multicolumn{2}{c}{\textbf{Sce-level(B)}} &
			\multicolumn{2}{c}{\textbf{Sce-level(T)}} & \multirow{2}{*}{\textbf{Overall}} \\
			\cmidrule(lr){2-3} \cmidrule(lr){4-5} \cmidrule(lr){6-7}
			& \textbf{IE} & \textbf{HR} & \textbf{TS} & \textbf{CS} & \textbf{TS} & \textbf{CS} & \\
			\midrule
			
			\multicolumn{8}{l}{\textcolor{gray}{\textit{Metrics - mIoU}}} \\
			\quad LISA-7B & 6.20 & 12.91 & 18.75 & 12.30 & 23.20 & 33.88 & 17.87 \\
			\quad LISA-13B & 10.24 & 15.96 & 19.97 & 14.22 & 30.64 & 33.77 & 20.80 \\
			\quad PixelLM-7B & 8.84 & 9.55 & 18.02 & 11.64 & 23.64 & 15.82 & 14.58 \\
			\quad PixelLM-13B & 11.36 & 14.42 & 20.22 & 12.29 & 34.88 & 32.44 & 20.94 \\
			\quad LISA-7B (FT) & \underline{13.35} & \underline{26.45} & \textbf{30.77} & \underline{28.76} & \underline{53.41} & \underline{44.06} & \underline{32.80} \\
			\quad GeoHeightChat & \textbf{16.03} & \textbf{31.23} & \underline{30.23} & \textbf{33.14} & \textbf{62.04} & \textbf{52.83} & \textbf{37.58} \\
			\midrule
			
			\multicolumn{8}{l}{\textcolor{gray}{\textit{Metrics - cIoU}}} \\
			\quad LISA-7B & 8.09 & 13.71 & 21.60 & 16.13 & 26.34 & 41.79 & 21.28 \\
			\quad LISA-13B & 11.62 & 17.45 & 21.99 & 20.34 & 33.73 & 40.86 & 24.34 \\
			\quad PixelLM-7B & 9.00 & 6.76 & 19.47 & 18.03 & 28.03 & 25.12 & 17.73 \\
			\quad PixelLM-13B & 11.86 & 13.52 & 24.46 & 19.70 & 36.76 & 39.56 & 24.31 \\
			\quad LISA-7B (FT) & \underline{18.81} & \underline{31.51} & \underline{44.23} & \underline{40.58} & \underline{60.72} & \underline{58.11} & \underline{42.33} \\
			\quad GeoHeightChat & \textbf{25.52} & \textbf{37.33} & \textbf{44.76} & \textbf{41.03} & \textbf{67.88} & \textbf{66.02} & \textbf{47.09} \\
			\bottomrule
		\end{tabular*}
	\end{table*}

	\begin{table*}[p]
		\centering
		\caption{Fine-grained mask generation on GeoHeight-Bench+, evaluated using mIoU (\%) and cIoU (\%). Scene-level results are reported separately for buildings (B) and trees (T).}\label{tab_4}
		\scriptsize
		\begin{tabular*}{\textwidth}{@{\extracolsep{\fill}} l cc cc cc cc c@{}}
			\toprule
			\multirow{2}{*}{\textbf{Model}} & \multicolumn{2}{c}{\textbf{Obj-level}} & \multicolumn{2}{c}{\textbf{Sce-level(B)}} &
			\multicolumn{2}{c}{\textbf{Sce-level(T)}} & \multicolumn{2}{c}{\textbf{Reason.-level}} & \multirow{2}{*}{\textbf{Overall}} \\
			\cmidrule(lr){2-3} \cmidrule(lr){4-5} \cmidrule(lr){6-7} \cmidrule(lr){8-9}
			& \textbf{IE} & \textbf{HR} & \textbf{TS} & \textbf{CS} & \textbf{TS} & \textbf{CS} & \textbf{LI} & \textbf{FI} & \\
			\midrule
			\multicolumn{10}{l}{\textcolor{gray}{\textit{Metrics - mIoU}}} \\
			\quad LISA-7B & 1.69 & 3.75 & 14.10 & 18.70 & 19.88 & 30.40 & 0.32 & 3.44 & 11.54 \\
			\quad LISA-13B & 2.49 & 4.48 & 12.73 & 20.63 & 17.98 & 28.18 & 1.32 & \underline{5.28} & 11.64 \\
			\quad PixelLM-7B & 1.26 & 3.11 & 12.40 & 19.01 & 18.87 & 24.02 & 12.16 & 0.01 & 11.36 \\
			\quad PixelLM-13B & 1.64 & 3.63 & 12.84 & 20.16 & 18.51 & 25.58 & 8.55 & 0.00 & 11.36 \\
			\quad LISA-7B (FT) & \underline{6.22} & \underline{8.29} & \underline{20.89} & \underline{28.65} & \underline{47.95} & \underline{40.04} & \underline{17.92} & 4.87 & \underline{21.85} \\
			\quad GeoHeightChat & \textbf{7.76} & \textbf{14.59} & \textbf{27.56} & \textbf{30.93} & \textbf{54.08} & \textbf{53.94} & \textbf{30.63} & \textbf{6.57} & \textbf{28.26} \\
			\midrule
			\multicolumn{10}{l}{\textcolor{gray}{\textit{Metrics - cIoU}}} \\
			\quad LISA-7B & 3.36 & 3.38 & 14.42 & 24.58 & 21.27 & 37.38 & 0.51 & 4.04 & 13.62 \\
			\quad LISA-13B & 3.66 & 4.32 & 12.09 & 29.27 & 18.67 & 34.50 & 1.68 & \underline{5.77} & 13.75 \\
			\quad PixelLM-7B & 1.81 & 1.92 & 12.12 & 18.11 & 19.64 & 25.16 & 10.74 & 0.01 & 11.19 \\
			\quad PixelLM-13B & 1.58 & 1.49 & 12.91 & 30.94 & 19.19 & 27.38 & 10.43 & 0.00 & 12.99 \\
			\quad LISA-7B (FT) & \underline{7.90} & \underline{10.03} & \underline{28.66} & \underline{43.44} & \underline{51.08} & \underline{58.22} & \underline{15.09} & 4.98 & \underline{27.43} \\
			\quad GeoHeightChat & \textbf{9.37} & \textbf{16.44} & \textbf{40.66} & \textbf{52.59} & \textbf{58.76} & \textbf{75.60} & \textbf{28.71} & \textbf{6.40} & \textbf{36.07} \\
			\bottomrule
		\end{tabular*}
	\end{table*}

	\subsection{Results}
	
	\label{sec:results}
	\noindent \textbf{Results on GeoHeight-Bench. }Because GeoHeightChat is
	fine-tuned on the benchmark's training split, the methodologically meaningful
	comparison is against the matched in-domain baseline LISA-7B~(FT), which is
	fine-tuned on the same data by directly concatenating optical, height, and
	categorical channels. Under this like-for-like setting (Table~\ref{tab_1} and
	Figure~\ref{fig:radar_text}, left), GeoHeightChat attains $44.14\%$ overall text
	accuracy against $31.73\%$ for LISA-7B~(FT), a gain of $+12.41$ points. Both models see the same training distribution, this comparison controls one
	important source of variation and is consistent with a benefit from the proposed
	implicit height-feature alignment; the ablations below provide additional evidence
	for this interpretation. The
	zero-shot closed- and open-source LMMs (e.g.\ GPT-4o $21.61\%$, Gemini-2.5-Flash
	$28.45\%$) are not like-for-like and are reported as zero-shot reference
	baselines that characterize how off-the-shelf models, without any task-specific
	adaptation, perform on height-aware tasks; the large margin over them therefore
	reflects in-domain fine-tuning as much as architectural design. The per-axis profile indicates that
	the advantage is broad rather than concentrated: relative to LISA-7B~(FT),
	GeoHeightChat improves on all seven task axes, most clearly on the
	height-dependent Pixel- and Object-level tasks (Elevation Retrieval $59.98\%$,
	Height Ranking $49.09\%$). The zero-shot baselines, by contrast, reveal a
	characteristic imbalance: LLaVA1.5-13B reaches a competitive $53.46\%$ on
	Elevation Retrieval yet collapses on Pixel Info., Instance Extraction, and Height
	Ranking ($\le 9.5\%$), showing that success on Elevation Retrieval alone does not
	transfer to broader height-related reasoning. The same pattern holds for fine-grained mask
	generation (Table~\ref{tab_3} and Figure~\ref{fig:radar_mask}, left), where
	GeoHeightChat reaches $37.58\%$ mIoU / $47.09\%$ cIoU versus $32.80\%$ / $42.33\%$
	for LISA-7B~(FT). Under matched data exposure, this gap is consistent with a
	benefit from aligning inferred height features rather than relying on naive
	channel concatenation.

	\noindent \textbf{Results on GeoHeight-Bench$+$. }We further evaluate on the more
	challenging GeoHeight-Bench+, which adds terrain attributes such as relief and
	slope. Against the matched LISA-7B~(FT) baseline (Table~\ref{tab_2} and
	Figure~\ref{fig:radar_text}, right), GeoHeightChat reaches $65.59\%$ overall
	versus $40.69\%$ ($+24.90$ points). The gain is largest at the Object level,
	where GeoHeightChat improves Instance Extraction from $48.74\%$ to $97.53\%$ and
	Height Ranking from $39.60\%$ to $94.87\%$; we note that these two tasks approach
	saturation, which warrants caution in over-interpreting the absolute values. For
	completeness we also list zero-shot models (e.g.\ the 30-billion-parameter
	LLaVA-NeXT-34B at $15.42\%$); as in the standard benchmark, these are zero-shot
	reference baselines rather than controlled comparisons. Crucially, the benefit of height
	priors is not uniform across tasks. On Slope Info. (SI), GeoHeightChat
	($18.58\%$) does not improve over and is in fact marginally below LISA-7B~(FT)
	($18.96\%$): unlike absolute height, slope and aspect are first-order spatial
	derivatives whose discriminative signal lies in the direction of the
	elevation gradient, which the distilled scalar height prior does not explicitly
	preserve. Recovering fine-grained gradient orientation from monocular optical
	cues thus remains an open problem that height alignment alone does not resolve.
	The segmentation results (Table~\ref{tab_4} and Figure~\ref{fig:radar_mask}, right) show the same uneven picture. On LI, which evaluates a terrain-based landslide-susceptibility proxy, implicit height geometry yields a substantial improvement, whereas pure optical baselines collapse to near zero (e.g.\ LISA-7B $0.32\%$ mIoU). On FI, which similarly represents a static terrain-based flood-susceptibility proxy, all methods perform poorly and the height prior brings little benefit: GeoHeightChat reaches only $6.57\%$ mIoU / $6.40\%$ cIoU, barely above LISA-7B~(FT). We attribute this to two factors: FI depends on absolute elevation together with hydrological context, such as drainage and depression connectivity, that monocular height priors do not capture, and its reference labels are derived from coarse DEM-depression heuristics, which bound the achievable accuracy. These results measure agreement with the benchmark's static susceptibility indicators and should not be interpreted as operational hazard-forecasting performance. Overall, GeoHeightChat attains $28.26\%$ mIoU / $36.07\%$ cIoU on GeoHeight-Bench+; the height prior clearly benefits object-level reasoning but offers limited or no improvement on derivative-gradient and terrain-based susceptibility tasks, which we identify as concrete limitations and directions for future work.

	\subsection{Cross-Dataset Generalization}
	\label{sec:vaihingen}
	To test generalization beyond the training distribution, and to provide
	indirect evidence against the concern that high in-domain scores might reflect
	memorized regional elevation priors rather than genuine height perception---we evaluate all models
	zero-shot on the ISPRS Vaihingen dataset, which is unseen
	during training and differs in sensor, resolution, and geographic region. Test
	patches are drawn by random cropping. Vaihingen provides reliable DEM/DTM
	products, so both the object-height (nDSM, GeoHeight) and terrain
	(DEM/DTM, GeoHeight+) reasoning tasks retain trustworthy ground truth.
	However, because the Vaihingen tiles are substantially smaller than those of GeoNRW, a randomly cropped patch spans only a narrow extent with little terrain relief. Consequently, the strongly terrain-dependent tasks---slope/aspect reasoning (SI) and the LI/FI susceptibility proxies---degenerate because their ground truth carries little discriminative signal within near-flat patches; we therefore exclude them.
	
	Tables~\ref{tab:vai_text}--\ref{tab:vai_mask} report the results. GeoHeightChat
	remains the strongest model on every aggregate under both settings: $29.83\%$
	versus $13.73\%$ overall text accuracy in the GeoHeight and GeoHeight+ settings
	respectively, each ahead of the best baseline (LLaVA-NeXT-13B, $18.85\%$; and
	GPT-4o, $14.77\%$), and it leads on mask generation with $37.41\%$ (Height) and
	$47.52\%$ (Elevation) overall cIoU. Two observations follow. First, in the
	terrain-conditioned GeoHeight+ setting, all evaluated models obtain lower overall
	text scores, although the magnitude of the decrease varies. This pattern is consistent with the weaker terrain signal
	available within small crops, but it does not by itself isolate the cause.
	Second, GeoHeightChat remains the top-ranked evaluated model on unseen geography,
	providing evidence that its advantage is not confined to the training regions.
	These results nevertheless do not rule out residual dataset-specific effects.
	The difficulty of terrain reasoning under limited
	spatial extent further motivates the explicit multi-sensor fusion discussed in
	Section~\ref{sec:limitations}.
	
	% ===== Table 1: GeoHeight text (nDSM) =====
	\begin{table*}[t]
		\centering
		\caption{Cross-dataset zero-shot text accuracy (\%) on ISPRS Vaihingen,
			GeoHeight setting (object-height / nDSM reasoning). Best in \textbf{bold}.}
		\label{tab:vai_text}
		\scriptsize
		\begin{tabular*}{\textwidth}{@{\extracolsep{\fill}} l cc cc ccc c @{}}
			\toprule
			\multirow{2}{*}{\textbf{Model}} & \multicolumn{2}{c}{\textbf{Pix-level}} & \multicolumn{2}{c}{\textbf{Obj-level}} & \multicolumn{3}{c}{\textbf{Sce-level}} & \multirow{2}{*}{\textbf{Overall}} \\
			\cmidrule(lr){2-3} \cmidrule(lr){4-5} \cmidrule(lr){6-8}
			& \textbf{ER} & \textbf{PI} & \textbf{IE} & \textbf{HR} & \textbf{PD} & \textbf{TS} & \textbf{CS} & \\
			\midrule
			GPT-4o            & 10.30 & 0.20 & 5.56 & 2.42 & 23.17 & 28.79 & 32.93 & 14.77 \\
			\midrule
			InternVL2-8B      & 11.52 & 3.43 & 7.64 & 6.15 & 32.64 & 19.39 & 34.75 & 16.50 \\
			LLaVA1.5-13B      & 46.06 & 4.24 & 2.78 & 2.42 & 16.23 & 36.36 & 13.94 & 17.43 \\
			LLaVA-NeXT-13B    & 42.42 & 6.87 & 3.47 & 3.42 & 25.96 & 35.45 & 14.34 & 18.85 \\
			\midrule
			GeoHeightChat     & \textbf{57.27} & \textbf{7.88} & \textbf{9.72} & \textbf{14.89} & \textbf{36.55} & \textbf{38.79} & \textbf{43.74} & \textbf{29.83} \\
			\bottomrule
		\end{tabular*}
	\end{table*}
	
	% ===== Table 2: GeoHeight+ text (DEM/DTM) =====
	\begin{table*}[t]
		\centering
		\caption{Cross-dataset zero-shot text accuracy (\%) on ISPRS Vaihingen,
			GeoHeight+ setting (terrain DEM/DTM reasoning).}
		\label{tab:vai_textplus}
		\scriptsize
		\begin{tabular*}{\textwidth}{@{\extracolsep{\fill}} l cc cc ccc c @{}}
			\toprule
			\multirow{2}{*}{\textbf{Model}} & \multicolumn{2}{c}{\textbf{Pix-level}} & \multicolumn{2}{c}{\textbf{Obj-level}} & \multicolumn{3}{c}{\textbf{Sce-level}} & \multirow{2}{*}{\textbf{Overall}} \\
			\cmidrule(lr){2-3} \cmidrule(lr){4-5} \cmidrule(lr){6-8}
			& \textbf{ER} & \textbf{PI} & \textbf{IE} & \textbf{HR} & \textbf{PD} & \textbf{TS} & \textbf{CS} & \\
			\midrule
			GPT-4o            & 1.84 & 0.00 & 0.69 & 2.45 & 13.93 & 19.39 & 24.14 & 8.92 \\
			\midrule
			InternVL2-8B      & 0.61 & 1.82 & 2.76 & 2.42 & 13.53 & 18.79 & \textbf{31.34} & 10.18 \\
			LLaVA1.5-13B      & 0.00 & 4.85 & \textbf{3.45} & 3.64 & 4.41 & 16.67 & 2.32 & 5.05 \\
			LLaVA-NeXT-13B    & 0.00 & \textbf{7.88} & 1.38 & 0.00 & 10.44 & 10.20 & 2.22 & 4.59 \\
			\midrule
			GeoHeightChat     & \textbf{7.27} & 6.06 & \textbf{3.45} & \textbf{8.48} & \textbf{16.21} & \textbf{25.76} & 28.89 & \textbf{13.73} \\
			\bottomrule
		\end{tabular*}
	\end{table*}
	
	% ===== Table 3: mask cIoU, Height(nDSM) + Elevation(DEM/DTM) =====
	\begin{table*}[t]
		\centering
		\caption{Cross-dataset zero-shot mask generation (cIoU) on ISPRS Vaihingen, split into building (B) and tree (T) targets.}
		\label{tab:vai_mask}
		\scriptsize
		\begin{tabular*}{\textwidth}{@{\extracolsep{\fill}} l cc cc cc c@{}}
			\toprule
			\multirow{2}{*}{\textbf{Model}} & \multicolumn{2}{c}{\textbf{Obj-level}} & \multicolumn{2}{c}{\textbf{Sce-level(B)}} & \multicolumn{2}{c}{\textbf{Sce-level(T)}} & \multirow{2}{*}{\textbf{Overall}} \\
			\cmidrule(lr){2-3} \cmidrule(lr){4-5} \cmidrule(lr){6-7}
			& \textbf{IE} & \textbf{HR} & \textbf{TS} & \textbf{CS} & \textbf{TS} & \textbf{CS} & \\
			\midrule
			\multicolumn{8}{@{}l}{\textcolor{gray}{\textit{Height (nDSM)}}} \\
			LISA-7B       & 15.09 & 29.51 & 27.53 & 30.87 & \textbf{34.48} & 54.89 & 32.06 \\
			LISA-13B      & \textbf{19.73} & 35.92 & 24.90 & 34.66 & 28.65 & 56.71 & 33.43 \\
			GeoHeightChat & 19.35 & \textbf{38.40} & \textbf{40.15} & \textbf{39.12} & 30.44 & \textbf{56.98} & \textbf{37.41} \\
			\midrule
			\multicolumn{8}{@{}l}{\textcolor{gray}{\textit{Elevation (DEM/DTM)}}} \\
			LISA-7B       & 12.81 & 25.69 & 39.92 & 37.93 & 18.84 & 58.95 & 32.36 \\
			LISA-13B      & 13.21 & 30.59 & 26.33 & 35.80 & 26.33 & 56.97 & 31.54 \\
			GeoHeightChat & \textbf{14.23} & \textbf{48.13} & \textbf{48.93} & \textbf{73.30} & \textbf{32.30} & \textbf{68.20} & \textbf{47.52} \\
			\bottomrule
		\end{tabular*}
	\end{table*}

	\begin{table*}[t]
		\centering
		\caption{Ablation study of the core designs on the GeoHeightChat. Here reports the text accuracy alongside mask generation metrics (mIoU and cIoU).}\label{tab_5}
		\scriptsize
		\begin{tabular*}{\textwidth}{@{\extracolsep{\fill}} l cc c ccc c @{}}
			\toprule
			\multirow{2}{*}{\textbf{Model Variants}} & \multicolumn{2}{c}{\textbf{Modality}} & \multirow{2}{*}{\textbf{Fusion Strategy}} & \multirow{2}{*}{\textbf{Text Acc.}} & \multicolumn{2}{c}{\textbf{Mask Acc.}} & \multirow{2}{*}{\textbf{Overall}} \\
			\cmidrule(lr){2-3} \cmidrule(lr){6-7}
			& \textbf{Height} & \textbf{Class} & & & mIoU & cIoU &  \\
			\midrule
			Baseline & \xmark & \xmark & - & -  & 17.87 & 21.28 & 19.58 \\
			\midrule
			Height Only & \cmark & \xmark & Adapt. Res. & 36.32 & \underline{30.95} & \underline{38.07} & \underline{35.11} \\
			Class Only & \xmark & \cmark & Adapt. Res. & 31.84 & 18.44 & 20.76 & 23.68 \\
			\midrule
			Concat Fusion & \cmark & \cmark & Concat. & \underline{40.87} & 28.58 & 32.30 & 33.92 \\
			GeoHeightChat & \cmark & \cmark & Adapt. Res. & \textbf{44.14} & \textbf{37.58} & \textbf{47.09} & \textbf{42.94} \\
			\bottomrule
		\end{tabular*}
	\end{table*}
	
	\subsection{Ablation Study}
	To validate the core designs of GeoHeightChat, we conduct ablation studies on GeoHeight-Bench (Table \ref{tab_5}), evaluating geometric supervision modalities (Stage-1) and feature fusion strategies (Stage-2).
	
	\vspace{5pt}
	
	\noindent \textbf{Impact of Teacher Supervision Modalities.}
	As shown in Table \ref{tab_5}, the baseline without geospatial alignment obtains an aggregate score of 19.58\%. Supervising the GeoAdapter using height information alone raises the score to 35.11\%, indicating that useful height-related structure can be inferred from RGB cues. Using semantic maps alone yields 23.68\%, suggesting that categorical boundaries provide less task-relevant information than height supervision in this setting. Combining both modalities produces the strongest aggregate result, supporting the complementary roles of height structure and semantic boundaries.
	
	\vspace{5pt}
	
	\noindent \textbf{Effectiveness of the Feature Fusion Strategy.}
	To integrate inferred geometric features while retaining the LMM's pre-trained visual representation, we compare our adaptive residual connection ($\mathcal{P}(V_{rgb}+\lambda\cdot\hat{Z}_{geo})$) with standard channel-wise concatenation ($\mathcal{P}([V_{rgb},\hat{Z}_{geo}])$). The adaptive approach outperforms concatenation on all three reported metrics, increasing the aggregate score from 33.92\% to 42.94\% and mIoU from 28.58\% to 37.58\%. The learnable scalar $\lambda$ controls the contribution of the geometric features and may help preserve the visual information required by downstream tasks.
	
	\vspace{5pt}
	
	\noindent \textbf{Choice of Feature Alignment Strategy.} To justify the two-stage training paradigm adopted in GeoHeightChat, we investigate alternative strategies for bridging the modality gap between optical imagery and geospatial information. Table~\ref{tab_9} presents a quantitative comparison of three distinct alignment approaches:
	
	\begin{itemize}
		\item[$\bullet$] \textbf{Strategy 1 (S1) -- Modality Dropout:} A pseudo-alignment approach where both RGB and geospatial inputs are processed, but the GeoEncoder branch is randomly dropped with a 50\% probability during training. This forces the model to occasionally rely solely on RGB imagery, attempting to implicitly learn height distributions through extreme data augmentation without an explicit alignment loss. Recent studies indicate that such naive modality dropout often exacerbates modality competition~\cite{wei2025boosting}, leading to suboptimal unimodal representation learning since the dropped modality is treated merely as noise.
		
		\item[$\bullet$] \textbf{Strategy 2 (S2) -- Single-Stage Joint Training:} An end-to-end multi-task approach where both optical and geospatial data are input simultaneously. The geometric alignment loss ($\mathcal{L}_{align}$) between the GeoEncoder and the Vision Encoder's projector is computed and optimized concurrently with the downstream auto-regressive text generation and mask segmentation losses. While intuitive, this joint formulation is known to suffer from gradient conflict (or multi-task interference)~\cite{liu2021conflict, ruffini2025gradient}. The dense, pixel-level geometric gradients and the sparse, high-level semantic gradients often point in diverging directions, causing destructive interference at the shared projector.
		
		\item[$\bullet$] \textbf{Strategy 3 (Ours) -- Two-Stage Distillation:} The decoupled paradigm used in our final model. It separates GeoAdapter pre-training with dense geometric supervision from subsequent LLM fine-tuning. This design reduces simultaneous optimization pressure on the shared representations and preserves the learned structural prior during instruction tuning.
	\end{itemize}

	\begin{table*}[t]
		\centering
		\caption{Choice of feature alignment strategies. Metrics include text accuracy and fine-grained mask generation (mIoU and cIoU).}\label{tab_9}
		\scriptsize
		\begin{tabular*}{0.6\textwidth}{@{\extracolsep{\fill}} c ccc c @{}}
			\toprule
			\multirow{2}{*}{\textbf{Align.}} & \multirow{2}{*}{\textbf{Text Acc.}} & \multicolumn{2}{c}{\textbf{Mask Acc.}} & \multirow{2}{*}{\textbf{Overall}} \\
			\cmidrule(lr){3-4}
			&  & mIoU & cIoU &  \\
			\midrule
			S1 & 23.71 & 18.88 & 22.75 & 21.78 \\
			S2 & 29.76 & 21.02 & 24.12 & 24.97 \\
			S3(Ours) & \textbf{44.14} & \textbf{37.58} & \textbf{47.09} & \textbf{42.94} \\
			\bottomrule
		\end{tabular*}
	\end{table*}

	As shown in Table~\ref{tab_9}, modality dropout (S1) obtains the lowest values on all three reported aggregate metrics. Without an explicit alignment objective, it provides a weaker training signal for learning a structural mapping from optical features.
	
	Joint training (S2) explicitly introduces the alignment objective alongside the downstream objectives, but it also underperforms the decoupled approach. One plausible explanation is multi-task interference: the projector is optimized simultaneously using dense geometric supervision and higher-level text and mask objectives, which may make the geometric mapping harder to learn. Because we do not directly measure gradient alignment, we treat this explanation as a hypothesis rather than a demonstrated mechanism.
	
	By separating geometric alignment from downstream instruction tuning, the two-stage strategy (S3) attains the strongest reported results. Freezing the aligned GeoAdapter during Stage~2 preserves its learned representation while the remaining trainable components adapt to text and mask generation.
	
	\subsection{Comparison with Explicit Height Mapping Pipeline}
	\label{sec:explicit_pipeline}
	
	To test whether an explicit height product can substitute for learned height-feature alignment, we retrain two monocular height estimators on GeoNRW and use their predicted nDSMs to construct two-stage prompting pipelines. The height estimators are TSENet~\cite{chen2025tsenet} and IM2ELEVATION~\cite{liu2020im2elevation}.
	
	\begin{table*}[t]
		\centering
		\caption{Comparison of explicit height mapping pipelines, and GeoHeightChat on the GeoNRW text tasks.}
		\label{tab:explicit_vs_implicit}
		\scriptsize
		\setlength{\tabcolsep}{3pt}
		\resizebox{\textwidth}{!}{%
			\begin{tabular}{@{} l l cc cc ccc c @{}}
				\toprule
				\multirow{2}{*}{Backbone} & \multirow{2}{*}{Height source} &
				\multicolumn{2}{c}{Pix.} & \multicolumn{2}{c}{Obj.} &
				\multicolumn{3}{c}{Scene} & \multirow{2}{*}{Overall} \\
				\cmidrule(lr){3-4} \cmidrule(lr){5-6} \cmidrule(lr){7-9}
				& & ER & PI & IE & HR & PD & TS & CS & \\
				\midrule
				\multicolumn{10}{@{}l}{\textcolor{gray}{\textit{LMMs}}} \\
				\multirow{3}{*}{GPT-4o}
				& RGB only & 19.59 & 2.13 & 11.84 & 8.76 & 26.73 & \underline{51.68} & 37.70 & 22.63 \\
				& TSENet & \underline{51.35} & \textbf{27.78} & 42.96 & \underline{37.96} & \textbf{53.13} & 40.25 & 26.00 & \underline{39.92} \\
				& IM2ELE & \underline{51.35} & \underline{26.74} & 42.22 & 34.08 & \underline{51.39} & 39.49 & 25.42 & 38.67 \\
				\addlinespace[2pt]
				\multirow{3}{*}{Gemini-2.5}
				& RGB only & 33.33 & 7.32 & 44.00 & 31.82 & 28.42 & 34.31 & \underline{45.42} & 32.09 \\
				& TSENet & 21.62 & 20.93 & \underline{47.41} & 26.64 & 50.29 & 35.34 & 32.86 & 33.58 \\
				& IM2ELE & 22.52 & 16.67 & 41.48 & 37.02 & 48.72 & 39.96 & 32.32 & 34.10 \\
				\midrule
				\multicolumn{10}{@{}l}{\textcolor{gray}{\textit{Implicit alignment}}} \\
				GeoHeightChat & Implicit prior & \textbf{65.85} & 22.11 & \textbf{56.39} & \textbf{79.96} & 29.66 & \textbf{72.41} & \textbf{84.60} & \textbf{58.71} \\
				\bottomrule
		\end{tabular}}
	\end{table*}
	
	Table~\ref{tab:explicit_vs_implicit} evaluates whether a conventional two-stage pipeline can recover height-aware reasoning without specialised multimodal alignment. Supplying a predicted nDSM improves GPT-4o from $22.63\%$ to $39.92\%$ and Gemini-2.5-Flash from $32.09\%$ to $34.10\%$ overall, confirming that explicitly rendered height products can provide useful auxiliary cues. Nevertheless, the gains are strongly backbone- and task-dependent, and the best explicit configuration remains $18.79$ points below GeoHeightChat. GeoHeightChat achieves the highest accuracy on five of the seven tasks. These results indicate that retraining the height estimators on the target domain and exposing their outputs at inference improves LMMs, but does not provide the consistent task-level transfer obtained by learning a height-aligned representation during training.
	
	%	\clearpage
	\subsection{Comparison with Few-Shot Prompting}
	\label{sec:few_shot}
	
	We evaluate GPT-4o and Gemini-2.5-Flash with two or four exemplars to test whether in-context demonstrations can mitigate task-interpretation and output-format errors. Every test query remains RGB-only.
	
	\begin{table*}[t]
		\centering
		\caption{Effect of few-shot prompting on the GeoNRW text tasks. Zero-shot results are reproduced from Table~\ref{tab:explicit_vs_implicit}; all test-time visual inputs are RGB-only.}
		\label{tab:few_shot}
		\scriptsize
		\setlength{\tabcolsep}{3pt}
		\resizebox{\textwidth}{!}{%
			\begin{tabular}{@{} l c cc cc ccc c @{}}
				\toprule
				\multirow{2}{*}{Model} & \multirow{2}{*}{Exemplars} &
				\multicolumn{2}{c}{Pix.} & \multicolumn{2}{c}{Obj.} &
				\multicolumn{3}{c}{Scene} & \multirow{2}{*}{Overall} \\
				\cmidrule(lr){3-4} \cmidrule(lr){5-6} \cmidrule(lr){7-9}
				& & ER & PI & IE & HR & PD & TS & CS & \\
				\midrule
				\multirow{3}{*}{GPT-4o}
				& zero-shot & 19.59 & 2.13 & 11.84 & 8.76 & 26.73 & \underline{51.68} & 37.70 & 22.63 \\
				& $k{=}2$ & 49.38 & \underline{33.72} & 38.52 & 53.81 & 52.23 & 41.44 & 40.51 & 44.23 \\
				& $k{=}4$ & \underline{52.25} & \textbf{33.98} & 50.37 & 52.14 & 51.37 & 45.35 & 32.96 & \underline{45.49} \\
				\addlinespace[2pt]
				\multirow{3}{*}{Gemini-2.5-Flash}
				& zero-shot & 33.33 & 7.32 & 44.00 & 31.82 & 28.42 & 34.31 & \underline{45.42} & 32.09 \\
				& $k{=}2$ & 37.84 & 23.13 & 51.11 & \underline{60.01} & \underline{54.52} & 46.05 & 41.30 & 44.85 \\
				& $k{=}4$ & 31.53 & 24.61 & \underline{56.30} & 57.61 & \textbf{54.53} & 39.38 & 40.29 & 43.46 \\
				\midrule
				GeoHeightChat & --- & \textbf{65.85} & 22.11 & \textbf{56.39} & \textbf{79.96} & 29.66 & \textbf{72.41} & \textbf{84.60} & \textbf{58.71} \\
				\bottomrule
		\end{tabular}}
	\end{table*}
	
	Table~\ref{tab:few_shot} shows that few-shot prompting substantially improves both general-purpose LMMs: GPT-4o reaches $45.49\%$ overall ($+22.86$), and Gemini-2.5-Flash reaches $44.85\%$ ($+12.76$). The improvement is not monotonic in $k$, however, indicating sensitivity to exemplar composition. GeoHeightChat remains $13.22$ points above the best few-shot result and leads five of the seven tasks. Few-shot prompting instead performs better on Pixel Info. and Pattern Description, suggesting that demonstrations improve response calibration, but do not replace a learned height-aligned representation for height-conditioned reasoning and localization.

	\section{Discussion}

	\begin{figure}[t]
		\centering
		\includegraphics[width=0.8\textwidth]{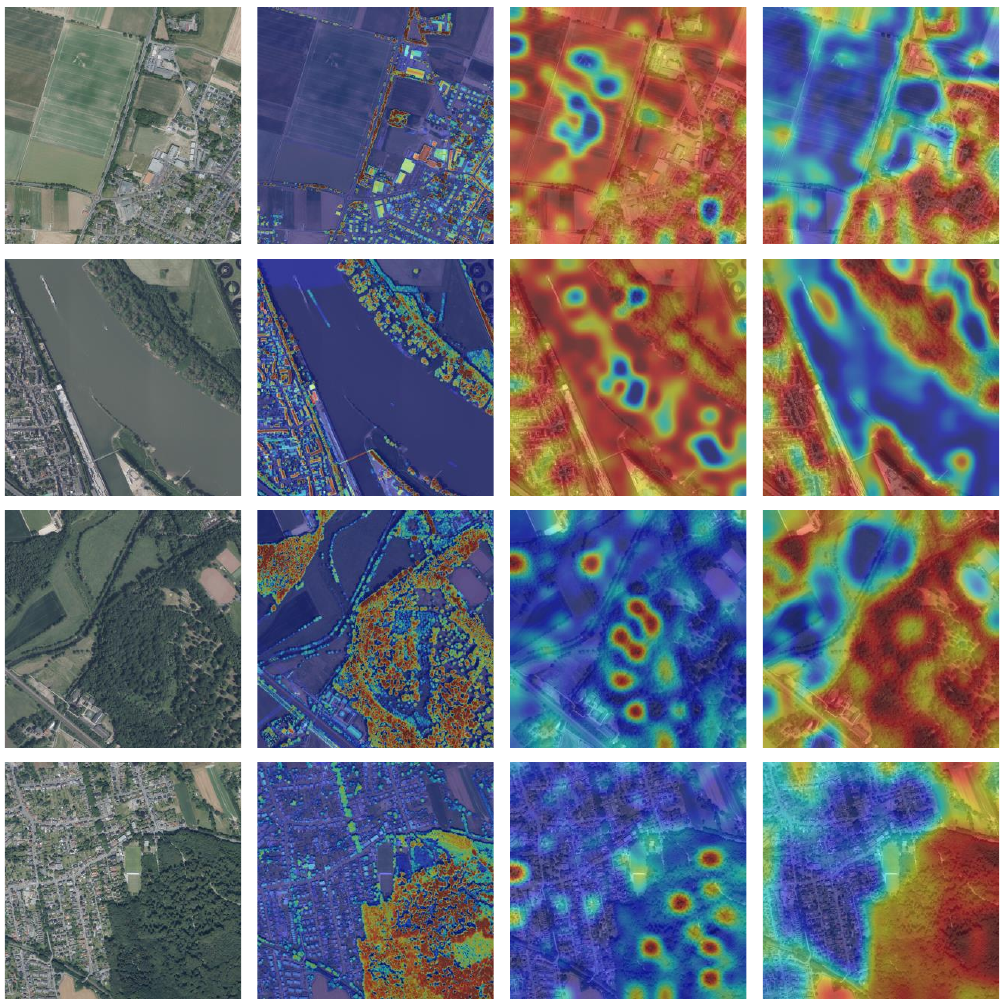}
		\caption{Each row is one GeoNRW tile. From left to right: the RGB input, the ground-truth nDSM, the PCA-1 pseudo-height of the raw CLIP features, and the PCA-1 pseudo-height of the GeoAdapter output $\hat{Z}_{geo}$; Raw CLIP activations are diffuse and weakly aligned with height, whereas the GeoAdapter recovers an nDSM-aligned height layout from the optical input alone.}
		\label{fig:clip_vs_geo}
	\end{figure}
	
	\begin{figure}[t]
		\centering
		\includegraphics[width=0.8\textwidth]{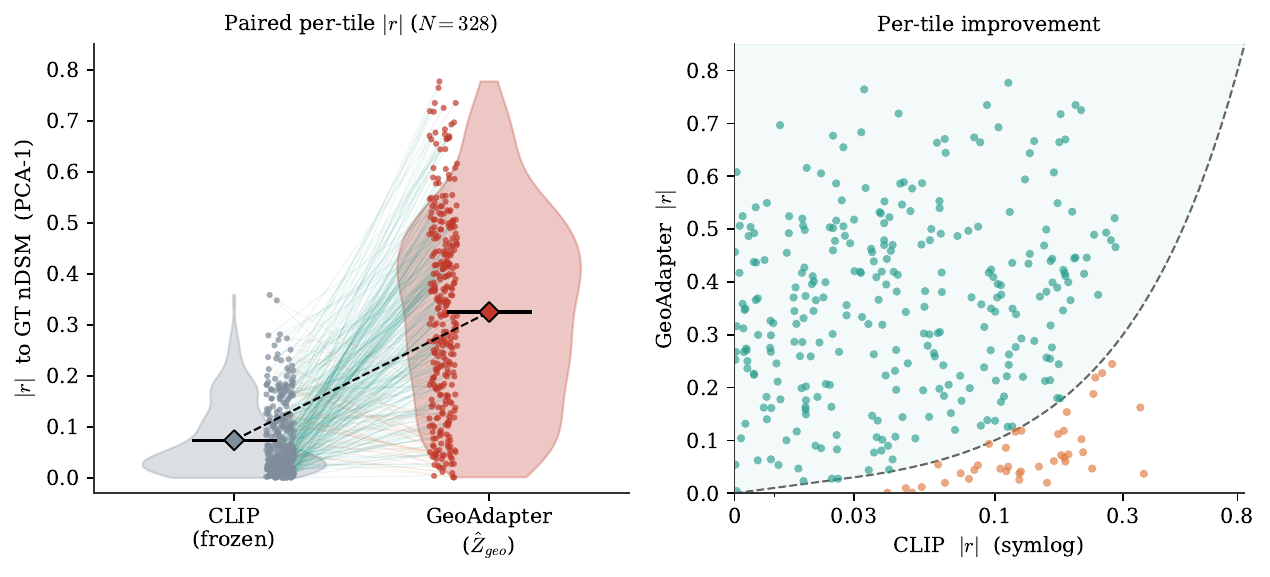}
		\caption{Per-tile absolute correlation $|r|$ between the PCA-1 pseudo-height of each feature map and the GeoNRW ground-truth nDSM. Left: paired distribution; each line links a tile's CLIP and GeoAdapter score (teal, improved; orange, worsened), and diamonds mark the means ($0.074\!\rightarrow\!0.325$). Right: the same pairs against the $y{=}x$ diagonal on a symlog $x$-axis; $88.1\%$ of tiles fall in the upper region.}
		\label{fig:clip_vs_geo_corr}
	\end{figure}
	
	\subsection{Task-wise Performance of Existing LMMs.}
	LMMs show relatively stronger results on several scene-level tasks (TS and CS) than on PI, IE, and HR, although some models also perform well on ER (Table~\ref{tab_1}). This uneven profile indicates that performance on conventional visual semantics does not reliably transfer to height-related localization and comparison. The segmentation results show a similar task dependence. Together, these observations motivate multimodal training data that explicitly represents height, while avoiding the stronger claim that optical LMMs contain no usable height cues.

	\subsection{The ``Vertical Blind Spot'' of Optical Models.} 
	Current models perform unevenly on height-related tasks. In Tables~\ref{tab_2} and \ref{tab_4}, several optical baselines (e.g., PixelLM) score near zero on LI and FI. As defined in Section~\ref{sec:benchmark_tasks}, these tasks represent static, terrain-based susceptibility indicators rather than complete landslide or flood hazard assessments. RGB imagery does not directly observe topography, although it may contain indirect cues correlated with terrain. GeoHeightChat improves substantially on the LI susceptibility proxy (e.g., 28.71\% cIoU), a result consistent with the value of height-related features for terrain-grounded applications. The benefit is task-dependent: performance on the FI proxy remains weak for all methods, including ours (Section~\ref{sec:results}), indicating that height features alone are insufficient and that relevant hydrological context is still missing.

	\subsection{Attribution of Height Awareness to GeoAdapter.}
	\label{sec:clip_probe}
	
	Does the alignment module make height-related structure more accessible than it is in the raw CLIP representation? We examine this question using a controlled linear diagnostic on the GeoNRW test set. For both the raw CLIP tokens $V_{rgb}$ and the adapter output $\hat{Z}_{geo}=\mathcal{A}(V_{rgb})$, we construct a scalar pseudo-height map using the same PCA-1 protocol and correlate it with the ground-truth nDSM. Because the adapter operates on the same CLIP features, the comparison measures how the adapter changes the linear accessibility of height-related structure under this diagnostic; it does not establish whether the backbone contains additional nonlinear height information.
	
	PCA-1 of the raw CLIP features is weakly aligned with nDSM ($\overline{|r|}=0.074$; Fig.~\ref{fig:clip_vs_geo}, third column). Passing the same features through the GeoAdapter raises the mean absolute correlation to $\overline{|r|}=0.325$, an increase of $\Delta\overline{|r|}{=}{+}0.251$ per tile. The increase occurs for $88.1\%$ of tiles and is statistically significant under a paired Wilcoxon test ($p\approx10^{-47}$, Fig.~\ref{fig:clip_vs_geo_corr}). Qualitatively, the adapter activations more clearly follow building clusters and vegetation boundaries in the nDSM (Fig.~\ref{fig:clip_vs_geo}, fourth column). These results support the narrower conclusion that the GeoAdapter makes height-related spatial structure more linearly accessible. Because $|r|$ measures spatial correspondence rather than metric elevation, the analysis does not demonstrate recovery of absolute physical scale.

	\subsection{Limitations and Future Directions}
	\label{sec:limitations}
	
	\noindent \textbf{Scope of the Terrain-Based Susceptibility Tasks.}
	LI and FI encode static susceptibility proxies derived from terrain and land-cover information. They do not incorporate the full set of variables required for operational hazard assessment, such as triggering rainfall, soil and geological conditions, drainage networks, flow connectivity, and temporal dynamics. The benchmark should therefore be interpreted as testing height-conditioned localization and reasoning under simplified terrain-based criteria, rather than forecasting the occurrence, magnitude, or impact of real hazard events.
	
	\vspace{5pt}
	
	\noindent \textbf{Absolute Scale in Monocular Reasoning and Multi-Sensor Fusion.} 
	While GeoHeightChat reduces the reliance on explicit height products by implicitly modeling geometric features, it still depends on monocular optical imagery during inference. Estimating absolute physical height from a single 2D top-down perspective is an inherently ill-posed problem, which can lead to over-reliance on planar visual semantics rather than an intrinsic grasp of absolute scale. A promising direction is therefore to move from implicit inference toward explicit multi-sensor fusion, for example pairing RGB with Light Detection and Ranging (LiDAR) or Synthetic Aperture Radar (SAR) \cite{yadav2025high}. This direction is in line with recent multi-sensor comprehension models and could enable more robust geometric reasoning for high-stakes expert systems such as disaster response and urban morphology analysis.
	
	\vspace{5pt}
	
	\noindent \textbf{Absence of Temporal Dynamics and Future Benchmarks.} 
	Currently, the GeoHeight-Bench $(+)$ evaluation framework predominantly focuses on static terrain morphology, relative height relationships, and stationary geospatial reasoning. It lacks the spatiotemporal dimensions necessary to capture the continuous evolution of geographical entities or dynamic disaster phenomena. In real-world applications, tracking volumetric changes over time, such as monitoring water level variations during floods, assessing post-earthquake structural collapse, or evaluating urban construction progress, is as critical as static height perception \cite{jakubik2023foundation}. Therefore, extending the current benchmark to encompass spatiotemporal dynamic tasks represents a vital direction for future work. By incorporating multi-temporal evaluations and leveraging advanced spatiotemporal foundation models, future frameworks can support more comprehensive, multidimensional reasoning for earth observation.

	\section{Conclusions}
	\label{section:E}
	
	In this work, we introduced GeoHeight-Bench to make height-aware reasoning in remote sensing LMMs measurable. Through a VLM-driven generation pipeline with human-in-the-loop verification, we constructed two benchmarks, GeoHeight-Bench and GeoHeight-Bench+, spanning pixel-, object-, scene-, and reasoning-level tasks. As a proof of concept, we further provided GeoHeightChat, a height-aware baseline that injects implicit vertical geometry into an optical LMM through two-stage training. Experiments show that current LMMs perform unevenly on height-related tasks, that aligning implicit height features yields gains on most evaluated height-dependent tasks, and that certain tasks, notably slope reasoning and terrain-based flood-susceptibility mapping, remain challenging. We hope GeoHeight-Bench serves as a reproducible testbed for the community and helps direct future work toward the vertical dimension of earth observation.
	
\clearpage

\appendix

\section{Visualizations of GeoHeight-Bench $(+)$ Samples}
\label{sec:B}

\begin{figure*}[h]
	\centering
	\includegraphics[width=\textwidth]{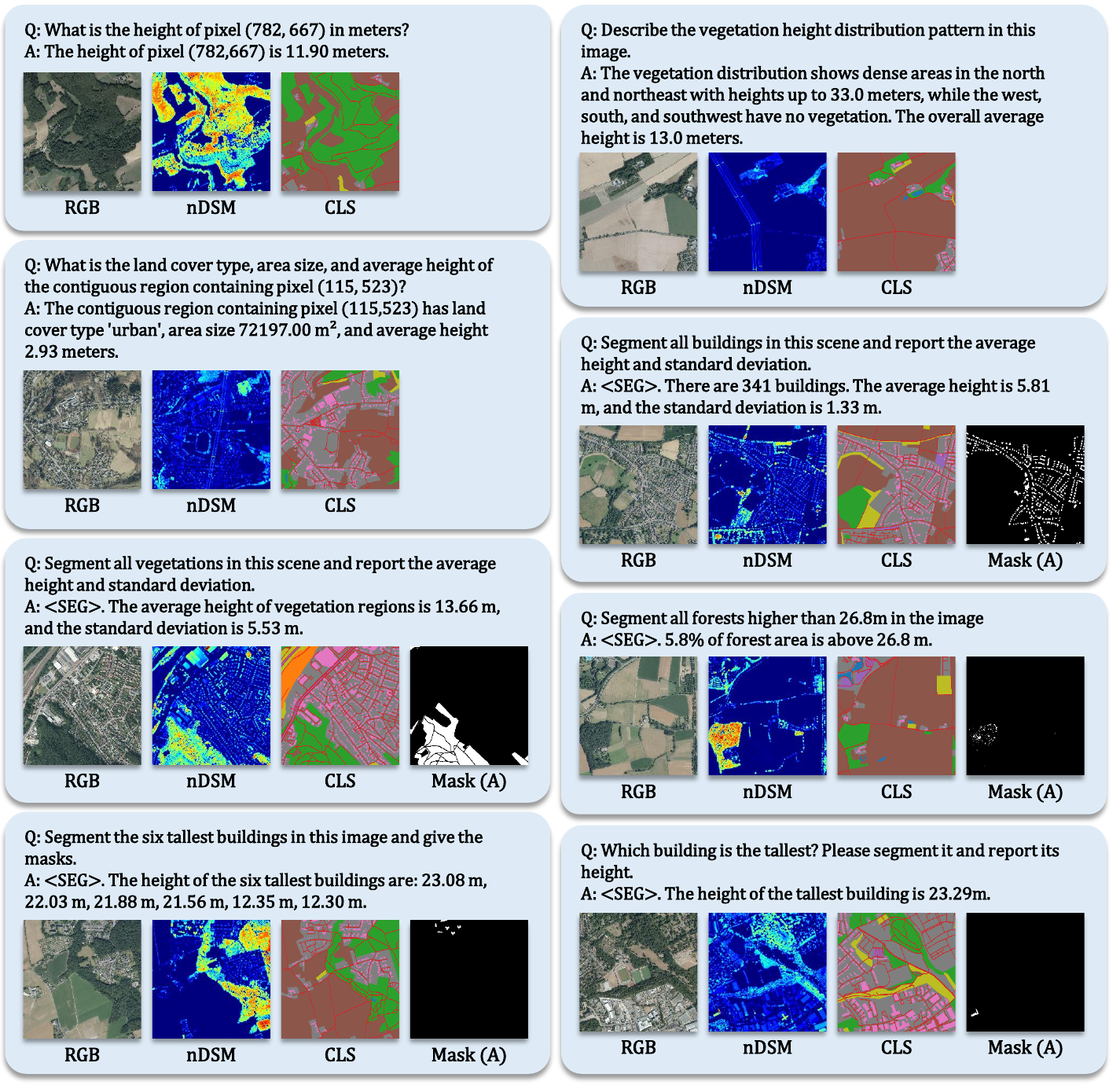}
	\caption{Additional examples from GeoHeight-Bench.}
	%	\label{fig_3}
\end{figure*}

\begin{figure*}[h]
	\centering
	\includegraphics[width=\textwidth]{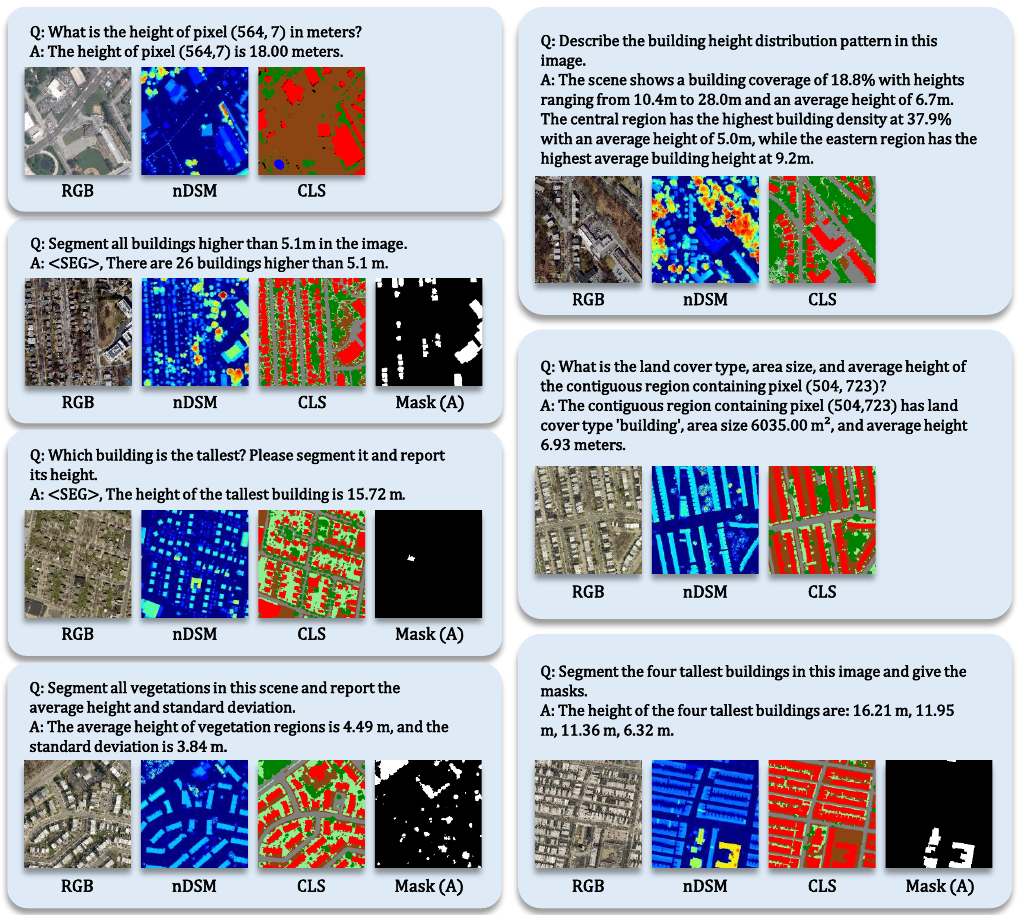}
	\caption{More samples of GeoHeight-Bench}
	%	\label{fig_3}
\end{figure*}

%\begin{figure*}[h]
%	\centering
%	\includegraphics[width=0.75\textwidth]{figures/GeoHeight-Bench_GeoNRW1.pdf}
%	\caption{More samples of GeoHeight-Bench}
%	%	\label{fig_3}
%\end{figure*}
%
%\begin{figure*}[h]
%	\centering
%	\includegraphics[width=0.8\textwidth]{figures/GeoHeight-Bench_GeoNRW2.pdf}
%	\caption{More samples of GeoHeight-Bench}
%	%	\label{fig_3}
%\end{figure*}
%
%\begin{figure*}[h]
%	\centering
%	\includegraphics[width=0.8\textwidth]{figures/GeoHeight-Bench_GeoNRW3.pdf}
%	\caption{More samples of GeoHeight-Bench}
%	%	\label{fig_3}
%\end{figure*}
%
%\begin{figure*}[h]
%	\centering
%	\includegraphics[width=0.8\textwidth]{figures/GeoHeight-Bench_RSMSS1.pdf}
%	\caption{More samples of GeoHeight-Bench}
%	%	\label{fig_3}
%\end{figure*}
%
%\begin{figure*}[h]
%	\centering
%	\includegraphics[width=0.8\textwidth]{figures/GeoHeight-Bench_RSMSS2.pdf}
%	\caption{More samples of GeoHeight-Bench}
%	%	\label{fig_3}
%\end{figure*}
%
%\begin{figure*}[h]
%	\centering
%	\includegraphics[width=0.8\textwidth]{figures/GeoHeight-Bench_RSMSS3.pdf}
%	\caption{More samples of GeoHeight-Bench}
%	%	\label{fig_3}
%\end{figure*}

\begin{figure*}[h]
	\centering
	\includegraphics[width=\textwidth]{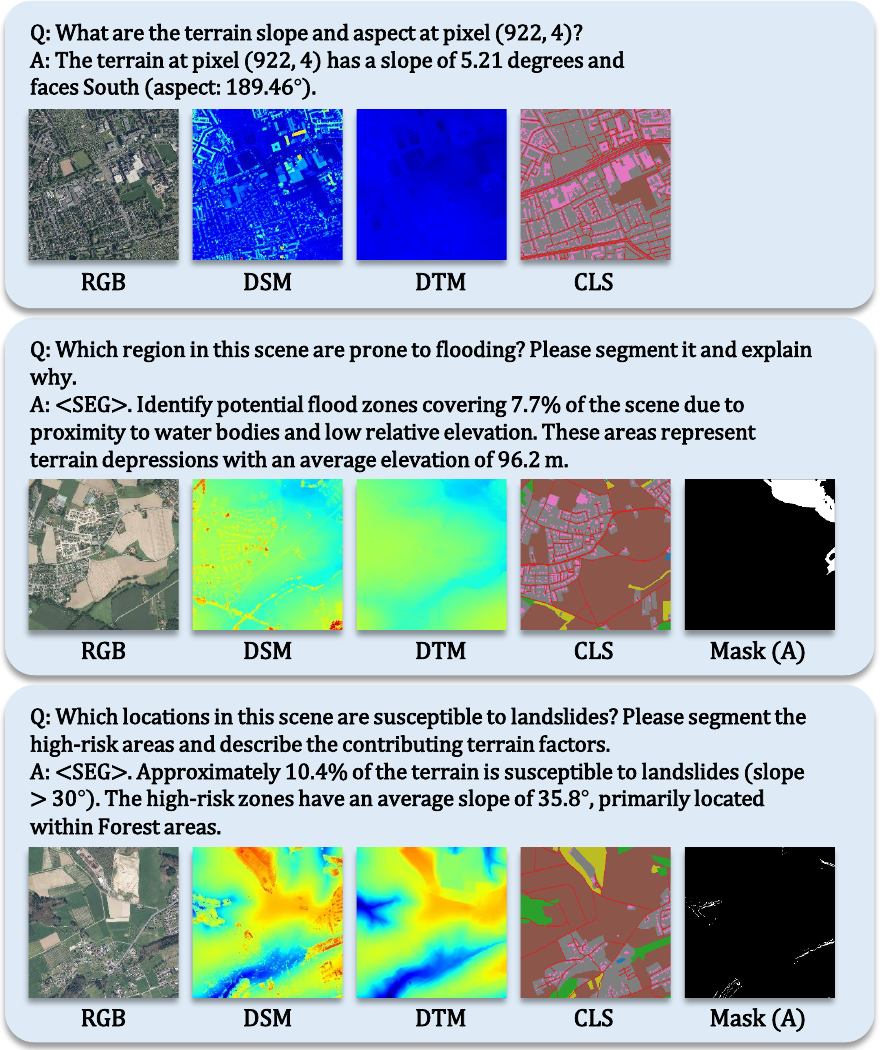}
	\caption{Additional examples from GeoHeight-Bench+.}
	%	\label{fig_3}
\end{figure*}

\clearpage

\section{Implementation Details of GeoHeightChat}
\label{sec:C}

\noindent \textbf{Model Architecture and Initialization.} 
We build GeoHeightChat upon the LISA-7B \cite{lai2024lisa} architecture. The vision backbone is initialized with the pre-trained \texttt{CLIP-ViT-Large-patch14} \cite{radford2021learning}, and the high-resolution decoder utilizes \texttt{SAM-ViT-H} \cite{kirillov2023segment}. Both the input RGB images and the geospatial target masks are processed at a resolution of $1024 \times 1024$. The maximum sequence length for the large language model is set to 512 tokens.

\vspace{5pt}

\noindent \textbf{Cross-Modal Geo-Alignment Modules.} 
For Stage-1 alignment, the GeoEncoder employs a frozen \texttt{ConvNeXt-Tiny} \cite{liu2022convnet} (ImageNet-initialized) to extract stable geometric targets from the geospatial inputs. The GeoAdapter processes RGB tokens using a 3-layer Bottleneck ResNet architecture. To guarantee training stability and prevent the catastrophic forgetting of visual semantics, we utilize \texttt{GroupNorm} and a Zero-Initialization strategy, initializing the residual scaling factor ($\gamma$) to zero so the blocks initially act as identity functions. A soft-clamp ($\pm 50.0$) is additionally applied to prevent numerical overflow during dense feature distillation.

\vspace{5pt}

\noindent \textbf{Training Setup for Instruction Tuning.} 
During the Geo-Aware Instruction Tuning phase, we load the pre-trained GeoAdapter (trained for 50 epochs in Stage-1) and strictly freeze its weights to preserve the learned RGB-to-Geometry alignment. The visual encoder and the multimodal projector are similarly frozen. To achieve parameter-efficient fine-tuning, we apply Low-Rank Adaptation \cite{hu2022lora} to the LLaMA-2-7B reasoning core, specifically targeting the query and value projection matrices (\texttt{q\_proj}, \texttt{v\_proj}). The LoRA hyperparameters are set to rank $r=8$, $\alpha=16$, with a dropout rate of $0.05$. The SAM mask decoder, language model head, and token embeddings remain fully trainable.

\vspace{5pt}

\noindent \textbf{Hyperparameters and Optimization.} 
The network is optimized using the AdamW optimizer with $\beta_1=0.9$, $\beta_2=0.95$, and zero weight decay. The base learning rate is set to $3 \times 10^{-4}$, regulated by a linear warmup schedule over the first 100 steps. The overall objective function integrates the text generation and mask segmentation losses with the following empirical weights: Cross-Entropy loss ($\lambda_{ce} = 1.0$), Mask Binary Cross-Entropy loss ($\lambda_{bce} = 2.0$), and Mask DICE loss ($\lambda_{dice} = 0.5$). 

\vspace{5pt}

\noindent \textbf{Hardware and Efficiency.} 
The training is conducted on a single compute node equipped with 8 NVIDIA RTX 4090 GPUs. To maximize training efficiency and accommodate high-resolution inputs, we utilize \texttt{bfloat16} mixed precision and DeepSpeed ZeRO \cite{rasley2020deepspeed} Stage-2 optimization with gradient checkpointing. The model is trained for 10 epochs (1,000 steps per epoch) with a per-device batch size of 1 and 10 gradient accumulation steps.

\newpage

\section{Details of Evaluation Metrics}
\label{sec:D}

To rigorously evaluate the diverse capabilities of Large Multimodal Models across varied geographic tasks, we implement a comprehensive evaluation protocol. Specifically, during the evaluation phase, we instruct the evaluated LMMs to generate responses following predefined templates. To accommodate minor format deviations and reliably extract the core answers, we employ Qwen2.5-1.5B as a universal information extractor (LLM-as-an-Extractor) \cite{zheng2023judging} to parse the generated text into structured numerical or categorical formats. Unless otherwise specified, all continuous value predictions are evaluated using a relative error threshold of $\tau = 20\%$ \cite{eigen2014depth}. The complete evaluation prompts, output-parsing prompts, and evaluation scripts will be released publicly together with the benchmark.

\subsection{Information Extraction Tasks}

\noindent \textbf{Continuous Value Accuracy}
For tasks involving continuous scalar predictions (e.g., elevation, height, area, count, and statistical metrics like mean $\mu$ and standard deviation $\sigma$), we evaluate using Threshold Accuracy. A predicted scalar $\hat{v}_i$ is deemed correct if its relative error concerning the ground truth $v_i$ is within $\tau=20\%$:
\begin{equation}
	Acc_{val} = \frac{1}{n} \sum_{i=1}^{n} \mathbb{I}\left( \frac{|v_i - \hat{v}_i|}{v_i} \le \tau\% \right)
\end{equation}
For zero-valued height or elevation targets ($v_i=0$), including the height assigned when the queried object is absent, a prediction is considered correct when its absolute error does not exceed $1\,\mathrm{m}$, i.e., $|\hat{v}_i-v_i|\leq1\,\mathrm{m}$.

\vspace{5pt}

\noindent \textbf{Instance and Sequence Evaluation}
For instance-level extraction (e.g., finding the tallest building) and sequence generation (e.g., top-$K$ height ranking), the LLM extractor parses outputs into a detection flag and a list of values: $(d, [h_1, \dots, h_m])$. The evaluation penalizes length inconsistencies (hallucinating extra objects or missing existing ones). For correctly detected instances, the accuracy is calculated by aligning the sorted predicted list with the ground truth and applying the relative threshold instance by instance.

\vspace{5pt}

\noindent \textbf{Topographic Aspect (Circular Variable)}
Unlike standard scalars, terrain aspect $\theta \in [0^\circ, 360^\circ)$ is a circular variable. To address the discontinuity at the $360^\circ/0^\circ$ boundary, we calculate the shortest circular difference: $\Delta \theta_i = \min(|\theta_i - \hat{\theta}_i|, 360^\circ - |\theta_i - \hat{\theta}_i|)$. A prediction is deemed correct if $\Delta \theta_i$ is within $20\%$ of the maximum possible angular error ($180^\circ$), yielding a strict threshold of $\Delta \theta_i \le 36^\circ$.

\vspace{5pt}

\noindent \textbf{Categorical \& Comprehensive Accuracy}
Categorical tasks (e.g., land cover classification) are evaluated via exact string matching aided by a domain-specific synonym dictionary. For tasks requiring both classification and continuous regression, we report a Comprehensive Accuracy computed as the unweighted macro-average of the categorical accuracy and the respective continuous accuracies.

\subsection{LLM-as-a-Judge for Open-Ended Descriptions}

For complex spatial distribution descriptions (e.g., vegetation or building layouts), standard string-matching metrics often fail to capture semantic nuances and appropriately penalize LMM hallucinations. Therefore, we adopt an LLM-as-a-Judge approach \cite{shabbir2025thinkgeo, schluckebier2025knowledge, canada2025multimodal} utilizing Qwen2.5-7B. 

The judge model evaluates the semantic and factual alignment between the prediction and the ground truth, assigning a discrete score $S_i \in \{1, 2, 3, 4, 5\}$. The specific grading rubric is defined as follows:
\begin{itemize}
	\item[$\bullet$] \textbf{Score 5 (Perfect Alignment):} The prediction flawlessly matches the ground truth, capturing all key statistics and spatial nuances.
	\item[$\bullet$] \textbf{Score 4 (Minor Deviations):} The prediction is generally accurate and factual but allows for minor numerical errors or slight omissions.
	\item[$\bullet$] \textbf{Score 3 (General Trend):} The prediction captures the overall layout or trend correctly but misses specific, detailed metrics.
	\item[$\bullet$] \textbf{Score 2 (Severe Hallucination):} The response is overly generic, vague, or contains significant hallucinated content.
	\item[$\bullet$] \textbf{Score 1 (Complete Contradiction):} The prediction entirely contradicts the ground truth or completely fails to address the prompt.
\end{itemize}

To unify this qualitative metric with our quantitative benchmarks, the discrete scores are normalized to a percentage-based accuracy:
\begin{equation}
	Acc_{desc} = \frac{1}{n} \sum_{i=1}^{n} \frac{S_i - 1}{4}
\end{equation}
where $n$ is the total number of evaluated descriptive samples.

%For complex spatial distribution descriptions (e.g., vegetation or building layouts), standard string-matching metrics fail to capture semantic nuances and penalize LMM hallucinations. Therefore, we adopt an LLM-as-a-Judge approach utilizing Qwen2.5-7B. 
%
%The judge model evaluates the semantic and factual alignment between the prediction and ground truth, assigning a discrete score $S_i \in \{1, 2, 3, 4, 5\}$. The rubric penalizes hallucinations and numerical inaccuracies: 5 denotes perfect alignment with all key statistics; 4 allows minor numerical deviations; 3 indicates capturing the general trend but missing specific metrics; 2 implies generic or severely hallucinated content; and 1 denotes complete contradiction. 
%
%To unify this qualitative metric with our quantitative benchmarks, the discrete scores are normalized to a percentage-based accuracy:
%\begin{equation}
%	Acc_{desc} = \frac{1}{n} \sum_{i=1}^{n} \frac{S_i - 1}{4}
%\end{equation}
%where $n$ is the total number of evaluated descriptive samples.

\newpage

\section{More Comparisons of GeoHeightChat and LMMs}
\label{sec:E}

\begin{figure*}[h]
	\centering
	\includegraphics[width=0.75\textwidth]{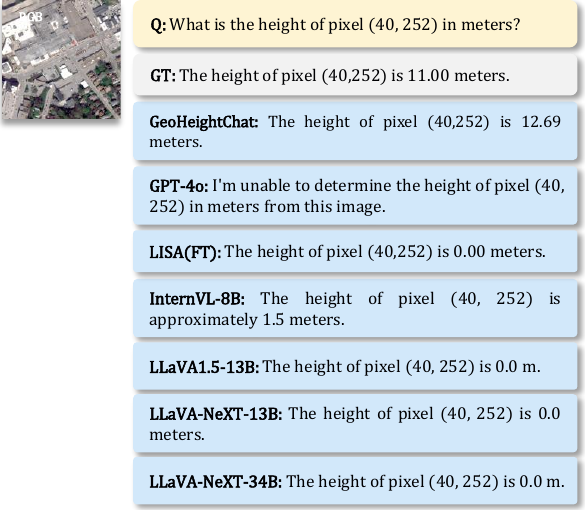}
	\caption{Qualitative comparison of model predictions with the ground truth for Elevation Retrieval (ER).}
	%	\label{fig_3}
\end{figure*}

\begin{figure*}[h]
	\centering
	\includegraphics[width=\textwidth]{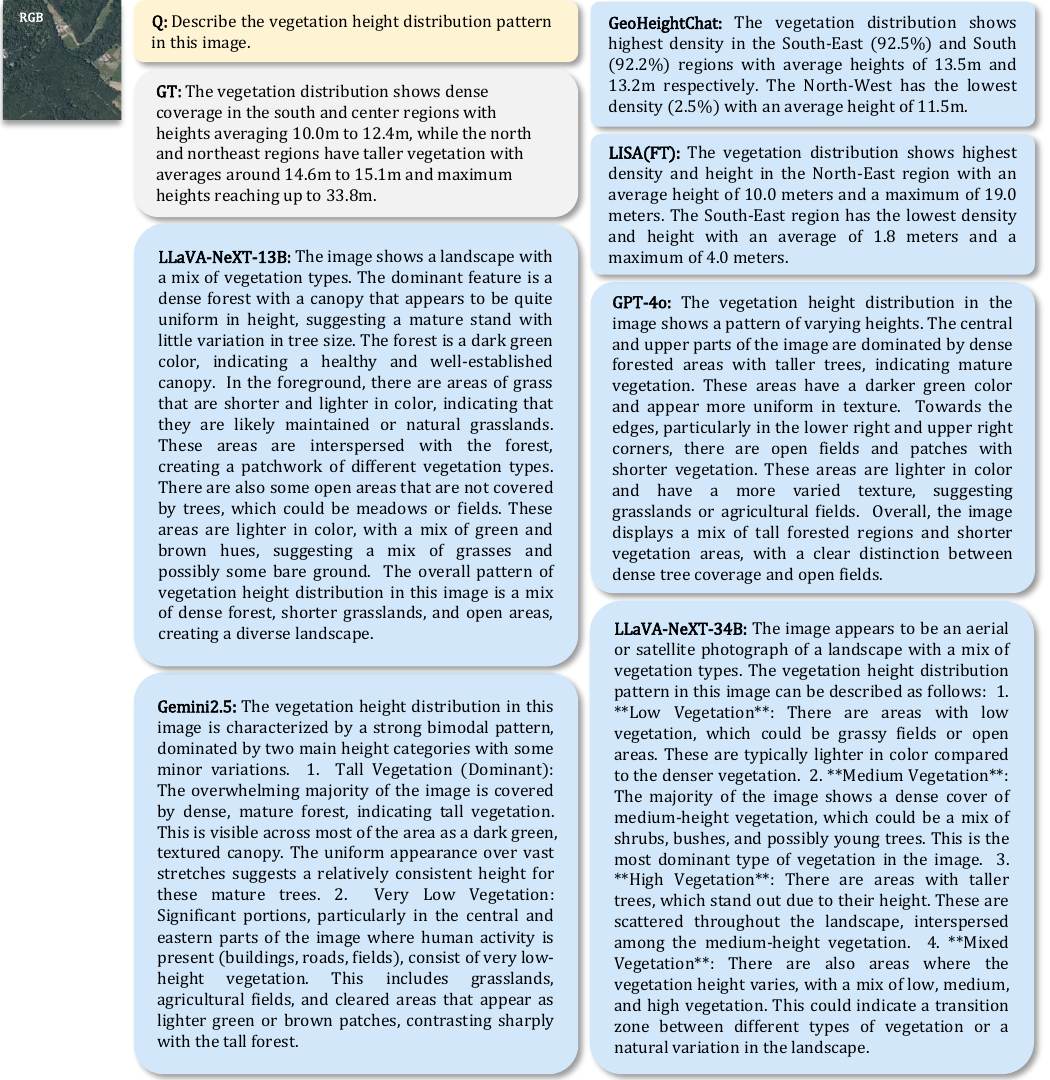}
	\caption{Qualitative comparison of model predictions with the ground truth for Pattern Description (PD).}
	%	\label{fig_3}
\end{figure*}

\clearpage

\begin{figure*}[h]
	\centering
	\includegraphics[width=\textwidth]{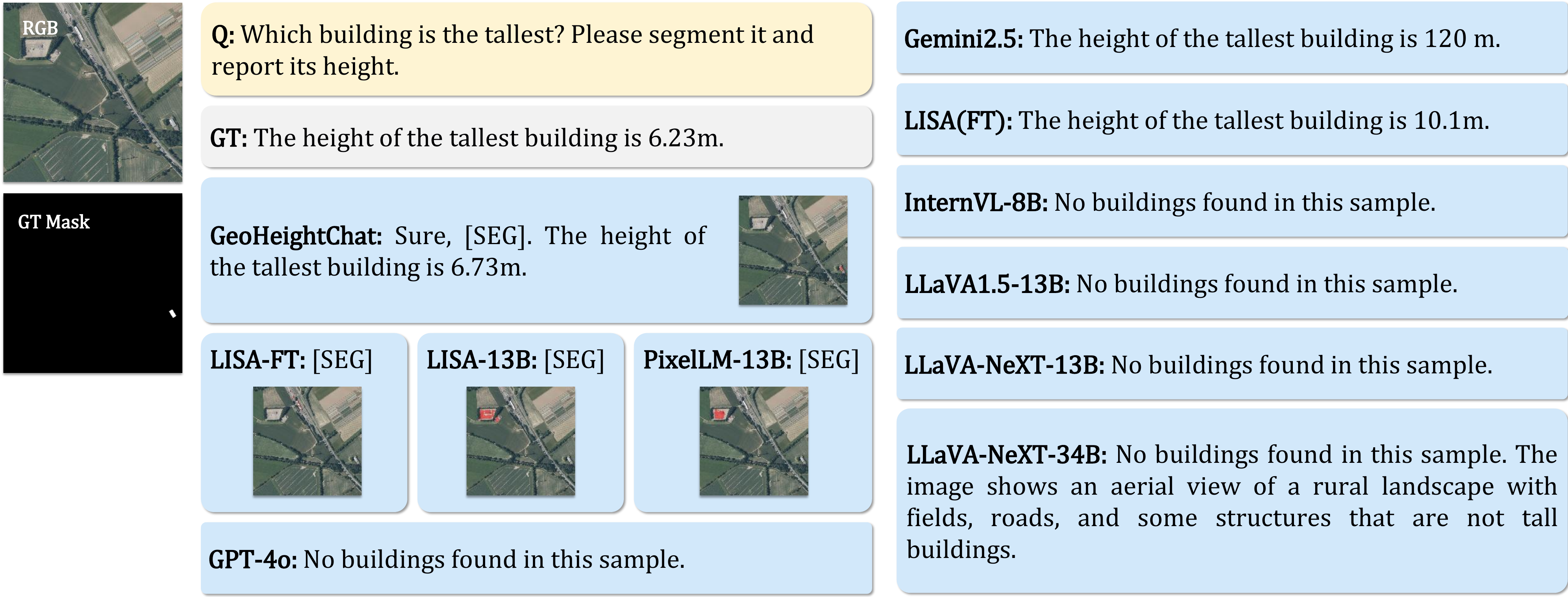}
	\caption{Qualitative comparison of model predictions with the ground truth for Instance Extraction (IE).}
	%	\label{fig_3}
\end{figure*}

\begin{figure*}[h]
	\centering
	\includegraphics[width=\textwidth]{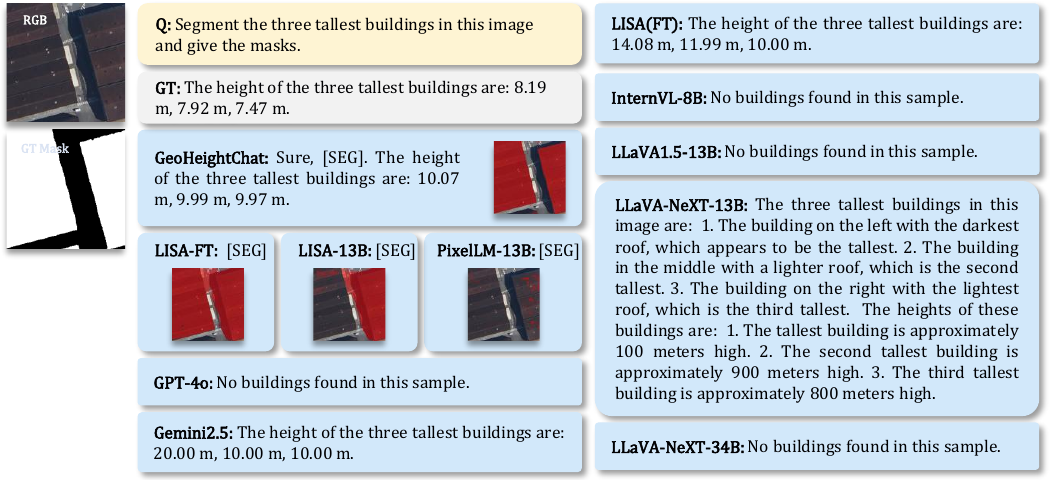}
	\caption{Qualitative comparison of model predictions with the ground truth for Height Ranking (HR).}
	%	\label{fig_3}
\end{figure*}

\clearpage

\section{Performance Breakdown by Source Dataset}
\label{sec:F}

Tables~\ref{tab_s1}--\ref{tab_s6} break down text-generation and mask-generation performance by the three source datasets: GeoNRW, FLAIR, and RSMSS. The comparison characterizes variation across geographic sources and spatial resolutions; because the subsets also differ in scene composition and label distribution, it should not be interpreted as a controlled estimate of any single dataset factor.

\begin{table*}[h]
	\centering
	\caption{Quantitative results for text generation and spatial reasoning on the GeoNRW subset of GeoHeight-Bench. Scores are reported as percentages.}\label{tab_s1}
	\scriptsize
%	\tiny
	\begin{tabular*}{\textwidth}{@{\extracolsep{\fill}} l cc cc ccc c @{}}
		\toprule
		\multirow{2}{*}{\textbf{Model}} & \multicolumn{2}{c}{\textbf{Pix-level}} & \multicolumn{2}{c}{\textbf{Obj-level}} & \multicolumn{3}{c}{\textbf{Sce-level}} & \multirow{2}{*}{\textbf{Overall}} \\
		\cmidrule(lr){2-3} \cmidrule(lr){4-5} \cmidrule(lr){6-8}
		& \textbf{ER} & \textbf{PI} & \textbf{IE} & \textbf{HR} & \textbf{PD} & \textbf{TS} & \textbf{CS} & \\
		\midrule
		
		\multicolumn{9}{l}{\textcolor{gray}{\textit{Closed-source LMMs}}} \\
		\quad GPT-4o & 19.59 & 2.13 & 11.84 & 8.76 & 26.73 & 51.68 & 37.70 & 22.63 \\
		\quad Gemini-2.5-Flash & 33.33 & 7.32 & \underline{44.00} & 31.82 & \underline{28.42} & 34.31 & 45.42 & 32.09 \\
		\midrule
		
		\multicolumn{9}{l}{\textcolor{gray}{\textit{Open-source LMMs}}} \\
		\quad InternVL2-2B & 13.11 & 5.30 & 23.05 & 8.25 & 20.43 & 43.75 & 15.04 & 18.42 \\
		\quad InternVL2-4B & 8.23 & 11.09 & 27.41 & 9.92 & 21.05 & 49.54 & 44.61 & 24.55 \\
		\quad InternVL2-8B & 17.38 & 1.83 & 20.87 & 8.04 & 20.04 & 27.74 & \underline{48.32} & 20.60 \\
		\quad LLaVA1.5-7B & 29.88 & 8.15 & 8.75 & 7.62 &  28.06 & 49.09 & 13.87 & 20.77 \\
		\quad LLaVA1.5-13B & 55.49 & 8.35 & 7.79 & 7.62 & 26.97 & 40.07 & 14.53 & 22.97 \\
		\quad LLaVA-NeXT-7B-Mistral & 10.67 & 0.00 & 7.79 & 9.76 & 17.95 & 49.09 & 14.79 & 15.72 \\
		\quad LLaVA-NeXT-7B-Vicuna & 24.09 & 6.17 & 9.03 & 8.54 & 15.95 & 49.16 & 13.82 & 18.11 \\
		\quad LLaVA-NeXT-13B & \underline{63.72} & 7.34 & 9.97 & 11.89 & 20.72 & 49.24 & 14.43 & 25.33 \\
		\quad LLaVA-NeXT-34B & 61.43 & 6.02 & 10.90 & 10.67 & 23.64 & 48.71 & 14.78 & 25.16 \\
		\quad Ministral3-8B & 15.68 & 5.75 & 4.67 & 6.10 & 16.55 & 36.15 & 10.59 & 13.64 \\
		\quad Pixtral-12B & 31.18 & 6.12 & 7.79 & 8.54 & 22.40 & 45.03 & 16.70 & 19.68 \\
		
		\midrule
		\multicolumn{9}{l}{\textcolor{gray}{\textit{Height-aware LMMs}}} \\
		\quad LISA-7B (FT) & \underline{63.72} & \underline{18.63} & 28.97 & \underline{35.07} & 12.22 & \underline{51.98} & 28.05 & \underline{34.09} \\
		\quad GeoHeightChat & \textbf{65.85} & \textbf{22.11} & \textbf{56.39} & \textbf{79.96} & \textbf{29.66} & \textbf{72.41} & \textbf{84.60} & \textbf{58.71} \\
		\bottomrule
	\end{tabular*}
\end{table*}

\begin{table*}[h]
	\centering
	\caption{Quantitative results for text generation and spatial reasoning on the FLAIR subset of GeoHeight-Bench. Scores are reported as percentages.}\label{tab_s2}
	\scriptsize
%	\tiny
	\begin{tabular*}{\textwidth}{@{\extracolsep{\fill}} l cc cc ccc c @{}}
		\toprule
		\multirow{2}{*}{\textbf{Model}} & \multicolumn{2}{c}{\textbf{Pix-level}} & \multicolumn{2}{c}{\textbf{Obj-level}} & \multicolumn{3}{c}{\textbf{Sce-level}} & \multirow{2}{*}{\textbf{Overall}} \\
		\cmidrule(lr){2-3} \cmidrule(lr){4-5} \cmidrule(lr){6-8}
		& \textbf{ER} & \textbf{PI} & \textbf{IE} & \textbf{HR} & \textbf{PD} & \textbf{TS} & \textbf{CS} & \\
		\midrule
		
		\multicolumn{9}{l}{\textcolor{gray}{\textit{Closed-source LMMs}}} \\
		\quad GPT-4o & 12.09 & 10.41 & 12.48 & 8.71 & 23.99 & 32.08 & 31.70 & 18.78 \\
		\quad Gemini-2.5-Flash & 26.56 & \underline{16.08} & \textbf{24.78} & \underline{17.11} & \textbf{27.98} & 35.00 & \textbf{36.99} & \underline{26.36} \\
		\midrule
		
		\multicolumn{9}{l}{\textcolor{gray}{\textit{Open-source LMMs}}} \\
		\quad InternVL2-2B & 24.64 & 8.73 & 8.80 & 0.32 & 10.40 & 32.58 & 10.68 & 13.74 \\
		\quad InternVL2-4B & 17.18 & 11.26 & 12.46 & 3.74 & 20.71 & \underline{37.00} & 22.76 & 17.87 \\
		\quad InternVL2-8B & 9.72 & 9.57 & 10.80 & 0.06 & 10.40 & 27.00 & 34.62 & 14.60 \\
		\quad LLaVA1.5-7B & 40.03 & 13.60 & 9.30 & 0.60 & \underline{25.06} & 32.75 & 10.73 & 18.87 \\
		\quad LLaVA1.5-13B & \underline{49.92} & 12.87 & 8.64 & 0.48 & 17.77 & 29.42 & 10.33 & 18.49 \\
		\quad LLaVA-NeXT-7B-Mistral & 27.07 & 5.92 & 8.97 & 0.32 & 20.59 & 33.42 & 11.79 & 15.44 \\
		\quad LLaVA-NeXT-7B-Vicuna & 42.43 & 12.68 & 9.63 & 0.32 & 15.92 & 32.83 & 10.71 & 17.79 \\
		\quad LLaVA-NeXT-13B & 48.21 & 10.77 & 9.65 & 0.55 & 20.10 & 33.33 & 10.52 & 19.02 \\
		\quad LLaVA-NeXT-34B & 42.32 & 5.83 & 11.31 & 0.80 & 19.21 & 36.75 & 12.32 & 18.36 \\
		\quad Ministral3-8B & 21.63 & 5.96 & 6.16 & 0.08 & 12.14 & 24.61 & 8.64 & 11.32 \\
		\quad Pixtral-12B & 31.98 & 8.24 & 12.48 & 0.42 & 17.90 & 32.59 & 11.59 & 16.46 \\
		
		\midrule
		\multicolumn{9}{l}{\textcolor{gray}{\textit{Height-aware LMMs}}} \\
		\quad LISA-7B (FT) & 49.76 & 15.34 & 17.61 & 15.71 & 17.29 & 35.67 & 30.72 & 26.01 \\
		\quad GeoHeightChat & \textbf{54.78} & \textbf{25.86} & \underline{20.43} & \textbf{27.27} & 21.11 & \textbf{38.00} & \underline{32.86} & \textbf{31.47} \\
		\bottomrule
	\end{tabular*}
\end{table*}

\begin{table*}[h]
	\centering
	\caption{Quantitative results for text generation and spatial reasoning on the RSMSS subset of GeoHeight-Bench. Scores are reported as percentages.}\label{tab_s3}
	\scriptsize
%	\tiny
	\begin{tabular*}{\textwidth}{@{\extracolsep{\fill}} l cc cc ccc c @{}}
		\toprule
		\multirow{2}{*}{\textbf{Model}} & \multicolumn{2}{c}{\textbf{Pix-level}} & \multicolumn{2}{c}{\textbf{Obj-level}} & \multicolumn{3}{c}{\textbf{Sce-level}} & \multirow{2}{*}{\textbf{Overall}} \\
		\cmidrule(lr){2-3} \cmidrule(lr){4-5} \cmidrule(lr){6-8}
		& \textbf{ER} & \textbf{PI} & \textbf{IE} & \textbf{HR} & \textbf{PD} & \textbf{TS} & \textbf{CS} & \\
		\midrule
		
		\multicolumn{9}{l}{\textcolor{gray}{\textit{Closed-source LMMs}}} \\
		\quad GPT-4o & 26.53 & 8.51 & 12.27 & 7.06 & 23.95 & 52.26 & 33.28 & 23.41 \\
		\quad Gemini-2.5-Flash & 24.44 & 10.94 & 20.69 & 18.78 & 24.94 & 42.59 & \textbf{45.82} & 26.89 \\
		\midrule
		
		\multicolumn{9}{l}{\textcolor{gray}{\textit{Open-source LMMs}}} \\
		\quad InternVL2-2B & 15.58 & 3.27 & 24.55 & 7.25 & 16.42 & 41.24 & 7.11 & 16.49 \\
		\quad InternVL2-4B & 11.26 & 8.45 & 26.14 & 9.59 & 23.94 & 52.78 & 34.75 & 23.84 \\
		\quad InternVL2-8B & 14.94 & 0.29 & 8.86 & 8.14 & 20.74 & 43.13 & 32.96 & 18.44 \\
		\quad LLaVA1.5-7B & 40.26 & 5.24 & 5.45 & 8.02 & \textbf{29.65} & 52.36 & 5.48 & 20.92 \\
		\quad LLaVA1.5-13B & 54.98 & 7.22 & 4.77 & 7.31 & 24.13 & 52.78 & 5.90 & 22.44 \\
		\quad LLaVA-NeXT-7B-Mistral & 30.59 & 0.00 & 4.77 & 7.25 & 19.44 & 52.47 & 5.71 & 17.18 \\
		\quad LLaVA-NeXT-7B-Vicuna & 35.68 & 8.64 & 10.23 & 8.63 & 22.32 & 51.64 & 5.84 & 20.43 \\
		\quad LLaVA-NeXT-13B & 54.98 & 10.32 & 8.18 & 9.56 & 20.38 & 52.38 & 6.30 & 23.16 \\
		\quad LLaVA-NeXT-34B & 50.36 & 11.70 & 9.09 & 8.12 & 17.65 & 52.07 & 5.91 & 22.13 \\
		\quad Ministral3-8B & 15.71 & 6.81 & 5.00 & 5.98 & 18.63 & 38.65 & 5.22 & 13.71 \\
		\quad Pixtral-12B & 28.65 & 8.94 & 8.86 & 7.61 & 20.20 & 43.28 & 8.40 & 17.99 \\
		
		\midrule
		\multicolumn{9}{l}{\textcolor{gray}{\textit{Height-aware LMMs}}} \\
		\quad LISA-7B (FT) & \underline{56.27} & \underline{18.63} & \underline{30.68} & \underline{27.98} & 18.85 & \underline{54.98} & 38.15 & \underline{35.08} \\
		\quad GeoHeightChat & \textbf{59.31} & \textbf{22.11} & \textbf{49.76} & \textbf{40.03} & \underline{27.71} & \textbf{55.30} & \underline{41.45} & \textbf{42.24} \\
		\bottomrule
	\end{tabular*}
\end{table*}

\vspace{1ex}
\noindent\textbf{1. Performance Variation Across Source Datasets.} GeoHeightChat attains the highest overall text score and mIoU on each source dataset. For cIoU, it leads on FLAIR and RSMSS, whereas LISA-7B~(FT) is higher on GeoNRW (40.27\% versus 38.80\%). GeoHeightChat's overall text score varies from 31.47\% on FLAIR to 58.71\% on GeoNRW. These differences may reflect spatial resolution, scene composition, geographic context, or label distribution; the present breakdown does not isolate their individual effects.

\vspace{1ex}
\noindent\textbf{2. Uneven Baseline Performance.} Across the three source datasets, LMMs are generally weaker on PI, IE, and HR than on several scene-level tasks, although some models attain high ER scores. This profile supports the claim that height-related performance is uneven across task types; it does not imply uniformly near-zero performance or establish that the models contain no height-related cues.

\vspace{1ex}
\noindent\textbf{3. Dense Prediction.} On FLAIR, GeoHeightChat achieves 49.32\% mIoU, compared with 45.13\% for the strongest matched baseline, LISA-7B~(FT), a gain of 4.19 points; the corresponding cIoU values are 55.22\% and 53.51\%. Improvements vary across source datasets, and the GeoNRW cIoU result is an exception. The aggregate pattern supports the usefulness of cross-modal geo-alignment for dense prediction without implying uniform superiority on every metric.

\begin{table*}[h]
	\centering
	\caption{Fine-grained mask generation on the GeoNRW subset of GeoHeight-Bench, evaluated using mIoU (\%) and cIoU (\%). Scene-level results are reported separately for buildings (B) and trees (T).}\label{tab_s4}
	\scriptsize
%	\tiny
	\begin{tabular*}{\textwidth}{@{\extracolsep{\fill}} l cc cc cc c@{}}
		\toprule
		\multirow{2}{*}{\textbf{Model}} & \multicolumn{2}{c}{\textbf{Obj-level}} & \multicolumn{2}{c}{\textbf{Sce-level(B)}} &
		\multicolumn{2}{c}{\textbf{Sce-level(T)}} & \multirow{2}{*}{\textbf{Overall}} \\
		\cmidrule(lr){2-3} \cmidrule(lr){4-5} \cmidrule(lr){6-7}
		& \textbf{IE} & \textbf{HR} & \textbf{TS} & \textbf{CS} & \textbf{TS} & \textbf{CS} & \\
		\midrule
		
		\multicolumn{8}{l}{\textcolor{gray}{\textit{Metrics - mIoU}}} \\
		\quad LISA-7B & 1.56 & 3.84 & 9.64 & 13.68 & 16.67 & 33.49 & 13.15 \\
		\quad LISA-13B & 2.60 & 5.45 & 9.15 & 17.77 & 24.28 & 32.79 & 15.34 \\
		\quad PixelLM-7B & 1.19 & 3.43 & 9.90 & 13.20 & 18.17 & 19.86 & 10.96 \\
		\quad PixelLM-13B & 1.72 & 3.98 & 8.31 & 14.05 & 21.13 & 31.83 & 13.50 \\
%		\quad LISAt-7B &  &  &  &  &  &  &  \\
		\quad LISA-7B (FT) & 3.71 & \underline{9.80} & \textbf{26.44} & \underline{33.19} & \underline{51.12} & \underline{49.31} & \underline{28.93} \\
		\quad GeoHeightChat & \textbf{5.49} & \textbf{11.60} & \underline{17.66} & \textbf{34.83} & \textbf{53.00} & \textbf{54.42} & \textbf{29.50} \\
		\midrule
		
		\multicolumn{8}{l}{\textcolor{gray}{\textit{Metrics - cIoU}}} \\
		\quad LISA-7B & 1.04 & 2.95 & 12.31 & 17.48 & 18.96 & 41.29 & 15.67 \\
		\quad LISA-13B & 2.94 & 6.21 & 10.72 & 24.78 & 25.49 & 40.57 & 18.45 \\
		\quad PixelLM-7B & 0.87 & 3.20 & 10.61 & 19.93 & 20.56 & 28.47 & 13.94 \\
		\quad PixelLM-13B & 1.32 & 2.73 & 11.47 & 22.37 & 22.20 & 35.98 & 16.01 \\
%		\quad LISAt-7B &  &  &  &  &  &  &  \\
		\quad LISA-7B (FT) & \underline{4.32} & \underline{11.85} & \textbf{46.75} & \textbf{51.07} & \underline{56.28} & \underline{71.33} & \textbf{40.27} \\
		\quad GeoHeightChat & \textbf{5.53} & \textbf{12.48} & \underline{35.47} & \underline{48.92} & \textbf{57.86} & \textbf{72.53} & \underline{38.80} \\
		\bottomrule
	\end{tabular*}
\end{table*}

\begin{table*}[h]
	\centering
	\caption{Fine-grained mask generation on the FLAIR subset of GeoHeight-Bench, evaluated using mIoU (\%) and cIoU (\%). Scene-level results are reported separately for buildings (B) and trees (T).}\label{tab_s5}
	\scriptsize
%	\tiny
	\begin{tabular*}{\textwidth}{@{\extracolsep{\fill}} l cc cc cc c@{}}
		\toprule
		\multirow{2}{*}{\textbf{Model}} & \multicolumn{2}{c}{\textbf{Obj-level}} & \multicolumn{2}{c}{\textbf{Sce-level(B)}} &
		\multicolumn{2}{c}{\textbf{Sce-level(T)}} & \multirow{2}{*}{\textbf{Overall}} \\
		\cmidrule(lr){2-3} \cmidrule(lr){4-5} \cmidrule(lr){6-7}
		& \textbf{IE} & \textbf{HR} & \textbf{TS} & \textbf{CS} & \textbf{TS} & \textbf{CS} & \\
		\midrule
		
		\multicolumn{8}{l}{\textcolor{gray}{\textit{Metrics - mIoU}}} \\
		\quad LISA-7B & 12.34 & 22.74 & 25.74 & 15.87 & 25.14 & 35.87 & 22.95 \\
		\quad LISA-13B & 19.98 & 28.08 & 28.72 & 17.30 & 35.86 & 36.18 & 27.69 \\
		\quad PixelLM-7B & 19.39 & 17.55 & 28.00 & 15.55 & 27.27 & 18.63 & 21.06 \\
		\quad PixelLM-13B & 23.28 & 26.76 & 34.74 & 17.62 & 46.35 & 39.23 & 31.33 \\
%		\quad LISAt-7B &  &  &  &  &  &  &  \\
		\quad LISA-7B (FT) & \underline{26.72} & \underline{50.75} & \underline{41.41} & \underline{34.70} & \underline{65.23} & \underline{51.97} & \underline{45.13} \\
		\quad GeoHeightChat & \textbf{29.84} & \textbf{57.36} & \textbf{44.27} & \textbf{34.84} & \textbf{68.70} & \textbf{60.92} & \textbf{49.32} \\
		\midrule
		
		\multicolumn{8}{l}{\textcolor{gray}{\textit{Metrics - cIoU}}} \\
		\quad LISA-7B & 13.85 & 23.57 & 29.17 & 19.62 & 30.71 & 39.31 & 26.04 \\
		\quad LISA-13B & 18.79 & 28.17 & 30.69 & 22.18 & 41.08 & 40.09 & 30.17 \\
		\quad PixelLM-7B & 15.92 & 10.21 & 29.06 & 20.97 & 33.30 & 28.19 & 22.94 \\
		\quad PixelLM-13B & 20.17 & 21.43 & 39.52 & 23.53 & 49.73 & 41.52 & 32.65 \\
%		\quad LISAt-7B &  &  &  &  &  &  &  \\
		\quad LISA-7B (FT) & \underline{36.82} & \underline{53.49} & \underline{45.92} & \textbf{46.96} & \underline{74.51} & \underline{63.33} & \underline{53.51} \\
		\quad GeoHeightChat & \textbf{39.21} & \textbf{58.08} & \textbf{48.53} & \underline{42.09} & \textbf{74.87} & \textbf{68.54} & \textbf{55.22} \\
		\bottomrule
	\end{tabular*}
\end{table*}

\begin{table*}[h]
	\centering
	\caption{Fine-grained mask generation on the RSMSS subset of GeoHeight-Bench, evaluated using mIoU (\%) and cIoU (\%). Scene-level results are reported separately for buildings (B) and trees (T).}\label{tab_s6}
	\scriptsize
%	\tiny
	\begin{tabular*}{\textwidth}{@{\extracolsep{\fill}} l cc cc cc c@{}}
		\toprule
		\multirow{2}{*}{\textbf{Model}} & \multicolumn{2}{c}{\textbf{Obj-level}} & \multicolumn{2}{c}{\textbf{Sce-level(B)}} &
		\multicolumn{2}{c}{\textbf{Sce-level(T)}} & \multirow{2}{*}{\textbf{Overall}} \\
		\cmidrule(lr){2-3} \cmidrule(lr){4-5} \cmidrule(lr){6-7}
		& \textbf{IE} & \textbf{HR} & \textbf{TS} & \textbf{CS} & \textbf{TS} & \textbf{CS} & \\
		\midrule
		
		\multicolumn{8}{l}{\textcolor{gray}{\textit{Metrics - mIoU}}} \\
		\quad LISA-7B & 4.69 & 12.16 & 20.87 & 7.36 & 27.80 & 32.27 & 17.53 \\
		\quad LISA-13B & 8.13 & 14.36 & 22.04 & 7.60 & 31.78 & \underline{32.34} & 19.38 \\
		\quad PixelLM-7B & 5.94 & 7.68 & 16.15 & 6.16 & 25.47 & 8.98 & 11.73 \\
		\quad PixelLM-13B & 9.07 & 12.51 & 17.62 & 5.21 & 37.17 & 26.26 & 17.97 \\
%		\quad LISAt-7B &  &  &  &  &  &  &  \\
		\quad LISA-7B (FT) & \underline{9.62} & \underline{18.79} & \underline{24.46} & \underline{18.39} & \underline{43.88} & 30.90 & \underline{24.34} \\
		\quad GeoHeightChat & \textbf{12.77} & \textbf{24.74} & \textbf{28.77} & \textbf{29.76} & \textbf{64.41} & \textbf{43.14} & \textbf{33.93} \\
		\midrule
		
		\multicolumn{8}{l}{\textcolor{gray}{\textit{Metrics - cIoU}}} \\
		\quad LISA-7B & 9.38 & 14.60 & 23.31 & 11.30 & 29.34 & \underline{44.77} & 22.12 \\
		\quad LISA-13B & 13.14 & 17.98 & 24.57 & 14.07 & 34.63 & 41.93 & 24.39 \\
		\quad PixelLM-7B & 10.20 & 6.88 & 18.73 & 13.18 & 30.24 & 18.70 & 16.32 \\
		\quad PixelLM-13B & 14.08 & 16.39 & 22.38 & 13.20 & 38.35 & 41.19 & 24.27 \\
%		\quad LISAt-7B &  &  &  &  &  &  &  \\
		\quad LISA-7B (FT) & \underline{15.30} & \underline{29.19} & \underline{40.02} & \underline{23.70} & \underline{51.38} & 39.66 & \underline{33.21} \\
		\quad GeoHeightChat & \textbf{31.83} & \textbf{41.43} & \textbf{50.28} & \textbf{32.09} & \textbf{70.92} & \textbf{56.98} & \textbf{47.26} \\
		\bottomrule
	\end{tabular*}
\end{table*}

\clearpage

	\bibliographystyle{elsarticle-num}
	\bibliography{reference}

\end{document}